\documentclass{article}
\usepackage[utf8]{inputenc}
\usepackage{amsmath,comment}
\usepackage{graphicx} 
\usepackage{amssymb}
\usepackage{dsfont}
\usepackage{float}
\usepackage[legalpaper, margin=1in]{geometry} 
\usepackage{soul,color}
\usepackage{hyperref}
\usepackage{todonotes}
\usepackage{physics}
\usepackage{natbib}

\newcommand{\floor}[1]{\left\lfloor #1 \right\rfloor}

\newcommand{\unbind}{{\setminus \!\!\! \circ \,}}
\newcommand{\hadarunbind}{{\;\setminus \!\!\!\! \odot \,}}
\newcommand{\ccunbind}{{\;\setminus \!\!\!\! \circledast \,}}
\newcommand{\lccunbind}{{\;\setminus \!\!\! \ast_{\!B} \,}}

\usepackage{tikz}

\title{
Computing on Functions Using Randomized Vector Representations
\thanks{
The work of DK was supported by the European Union's Horizon 2020 Programme under the Marie Skłodowska-Curie Individual Fellowship Grant (839179). 
The work of CJK was supported by the Department of Defense (DoD) through the National Defense Science \& Engineering Graduate (NDSEG) Fellowship Program.
FTS was supported by Intel and NIH R01-EB026955.
The work of BAO and DK  was supported in part by the DARPA's VIP (Super-HD Project) and AIE (HyDDENN Project) programs and by AFOSR FA9550-19-1-0241.
The work of FTS, BAO, and DK  was supported in part by Intel's THWAI program.
}
}
\author{E.~Paxon~Frady$^{1,2}$,  Denis~Kleyko$^{2,3}$, Christopher J. Kymn$^2$,\\ Bruno~A.~Olshausen$^2$, Friedrich~T.~Sommer$^{1,2}$ }
\date{1. Intel, Neuromorphic Computing Lab \\
2. UC Berkeley, Redwood Center for Theoretical Neuroscience\\
3. Research Institutes of Sweden, Intelligent Systems Lab\\

}

\begin{document}

\maketitle

\begin{abstract}
Vector space models for symbolic processing that encode symbols by random vectors have been proposed in cognitive science and connectionist communities under the names Vector Symbolic Architecture (VSA), and, synonymously, Hyperdimensional (HD) computing. In this paper, we generalize VSAs to function spaces by mapping continuous-valued data into a vector space such that the inner product between the representations of any two data points represents a similarity kernel. By analogy to VSA, we call this new function encoding and computing framework Vector Function Architecture (VFA). In VFAs, vectors can represent individual data points as well as elements of a function space (a reproducing kernel Hilbert space).  The algebraic vector operations, inherited from VSA, correspond to well-defined operations in function space.  Furthermore, we study a previously proposed method for encoding continuous data, fractional power encoding (FPE), which uses exponentiation of a random base vector to produce randomized representations of data points and fulfills the kernel properties for inducing a VFA. We show that the distribution from which elements of the base vector are sampled determines the shape of the FPE kernel, which in turn induces a VFA for computing with band-limited functions. In particular, VFAs provide an algebraic framework for implementing large-scale kernel machines with random features, extending \citet{rahimi2007random}.  Finally, we demonstrate several applications of VFA models to problems in image recognition, density estimation and nonlinear regression.  Our analyses and results suggest that VFAs constitute a powerful new framework for representing and manipulating functions in distributed neural systems, with myriad applications in artificial intelligence.

\end{abstract}

\section{Introduction}
The impressive recent achievements in artificial intelligence (AI) are driven by models that compute on data representations embedded in high-dimensional vector spaces, to date most notably, neural networks~\citep{bengio2021deep}. Once provided with input data and, if applicable, supervisory signals, neural networks provide powerful end-to-end training mechanisms to optimize performance for a given task.  However, state-of-the-art neural networks still suffer from two fundamental problems. One is that the application of learned knowledge is brittle and does not easily generalize to new contexts outside the training set. It has been argued that this problem is due in part to the inability to do variable binding \citep{fodor1988connectionism, greff2020binding}, which is crucial for enabling knowledge learned in one domain to be detached and flexibly applied to other domains \citep{Smolensky1990}. A second inherent problem with current neural networks is a lack of transparency. That is, both the vector representations and transformations in the neural network have no clear interpretation in terms of computation. Although the explicit computations whereby signals are combined through weighted sums and thresholding are well specified and understood for each neuron in the network, the overall system is often treated as a black box without gaining insight into the mathematical structure of the representations learned or the underlying computational function that is being performed by the network as a whole. The lack of transparency is an obstacle both in using neural networks as explanatory models for brain function, and in analyzing, understanding, or explaining the decisions of a neural network in specific application settings.

In recent years, we and others have made progress in advancing a vector space computing framework called Vector Symbolic Architecture (VSA), or synonymously Hyperdimensional (HD) computing, that both enables variable binding and is fully transparent \citep{Plate1994, Kanerva1996, gayler1998multiplicative, eliasmith2001integrating,RachkovskijBinding2001}.  In VSA, symbols, data, or other entities are represented by randomly mapping them into a vector space of fixed dimensionality.  An algebra over these vectors consisting of addition, multiplication and permutation enables the operations of bundling, binding and sequencing, respectively~\citep{KleykoComputingParadigm2021}.  There are now multiple successful examples of this approach applied to text analysis \citep{JonesMeaning2007, Joshi2016,
RecchiaSemantic2015}, decoding of EEG and EMG signals \citep{RahimiBiosignal2019,MoinWearable2021}, sequence learning \citep{HannaganHolographic2011,Frady2018}, and robotics \citep{neubert2019introduction}.  However, to date most of these applications have been limited to discrete data such as text, words or other tokens, or by discretizing data that are fundamentally continuous in nature and thereby ignoring important topological similarity relationships in the data \citep{edelman1998representation}.

In this paper, we turn to the question of how continuous data and functions may be represented in a vector space, and how they may be manipulated via the algebra of VSA.  We develop a new framework for computing with functions in a vector space, which by analogy to VSA we call Vector Function Architecture (VFA).  Like VSA, VFA is fully transparent.  Vectors can represent individual data points as well as elements of a function space that is well-defined as a reproducing kernel Hilbert space.  The domain of the functions can encode continuous-valued quantities in data, such as position, time or wavelength. 
Functions are manipulated by the vector operations of VSA:  addition of vectors corresponds to addition of functions, and binding of vectors corresponds to convolution of functions. 
The function domain is encoded by exponentiating a fixed random base vector — this encoding has been introduced before as fractional power encoding \citep{PlateRecurrent1992, Plate1994}.  Together, these encodings and operations open the door to a powerful new way of computing on data such as images, sound waveforms, and myriad other types of continuous information streams where one desires to manipulate functions — as is currently done in symbolic computing — but where information is represented holistically by elements of the system as in neural networks. 

Beyond its relation to VSA, VFA may be seen as part of a larger class of algorithms that leverage randomness to either expand dimensionality — e.g., hashing \citep{cormen2009introduction,dasgupta2008algorithms}, stochastic computing \citep{alaghi2018computing}, reservoir computing \citep{Frady2018,cuchiero2021discrete} — or reduce dimensionality — e.g., compressed sensing \citep{candes2006compressive,donoho2006compressed, frady2020variable}.  
In particular, VFAs provides an algebraic framework for implementing large-scale kernel machines with random features, including and extending \citet{rahimi2007random}.
On the other hand, what sets VFA apart from all of these methods is that it contains a mathematically grounded framework for computing on the high-dimensional representations.

We envision that the VFA framework will help towards building a type of AI which uses representations that are distributed, interpretable, and can be processed by algebraic vector operations with well-defined computational meaning. This AI approach could combine advantages of neural networks, probabilistic reasoning and symbolic AI; that is, the ability to learn, to be executable on distributed hardware, and to have the capability of rule-based reasoning for generalizing and extrapolating prior knowledge.  We believe the approach may also help provide insight into highly distributed representations in the brain, such as in entorhinal cortex and hippocampus.

The rest of the paper is organized as follows. Section~\ref{sec:background} revisits classical VSA models and states results from functional analysis that are relevant for this paper.
In Section~\ref{sec:VFA}, we reformulate symbolic VSA in terms of inner product kernels,  and generalize these kernels to describe the similarity of real-valued data mapped to vectors by locality-preserving encoding (LPE) . Under certain conditions, the resulting vector space model exhibits transparency and forms a VFA, i.e., the vectors represent functions of a defined function space and the VSA vector operations perform specific operations in the function space. 
Section~\ref{sec:VFA:FPE} describes the construction of VFA with an existing LPE method called fractional power encoding, or fractional binding \citep{PlateRecurrent1992, PlateAnalogical1994}. We generalize fractional power encoding to yield VFAs with phasor-valued, real-valued, and sparse representations, together with algebraic operations that can be efficiently implemented in hardware. In Section~\ref{sec:FPE:kernels}, we analyze the possible kernel shapes in VFAs. Notably, VFAs with kernels for a variety of different applications can be easily constructed by sampling the base vector in the FPE non-uniformly. In Section~\ref{sec:VFA:detection}, we describe how a given vector in VFA can be decoded and denoised. Decoding provides the transparency of VFA computations. Denoising,  similar as in VSAs, is essential for preventing error accumulation known from analog computers.  In Section~\ref{sec:kernel:methods}, we demonstrate applications of VFAs. Finally, the discussion in Section \ref{sec:discussion} summarizes our results, and explains implications for technical applications and neuroscience, as well as relations to the previous literature.



\section{Background}
\label{sec:background}

 \subsection{Symbolic VSA models}
\label{sec:background:symbolic:VSA}
The question of how to represent and process information in high-dimensional vector spaces has a long tradition not only in cognitive neuroscience \citep{GaylerJackendoff2003,gunther2019vector}, but also in language processing \citep{vanrijsbergen2004geometry, widdows2004geometry}. 
Here our starting point is vector models for symbolic cognitive reasoning, known as  Vector Symbolic Architectures (VSA) or Hyperdimensional Computing  \citep{plate1995holographic,Kanerva1997,gayler1998multiplicative,Rachkovskij2001}.

VSAs are models for symbolic reasoning with vector representations that have two distinguishing properties. The first is that symbols are represented by randomized $n$-dimensional vectors. 
In the classical VSA models, symbols are encoded by vectors whose components are drawn i.i.d. from a specified distribution. This encoding strategy ensures that representations of different symbols are well separated, where the \emph{similarity} of symbols is typically measured by the inner product between vectors, or a simple function of the inner product, such as cosine similarity. Here, we generally use the inner product with a model-dependent normalization factor to represent the similarity. 
The normalization factor will be omitted in the following for notational simplicity.
In VSA, a vector is decoded by comparing it with representation vectors of symbols stored in a codebook. Decoding and memory-based denoising is an important step in the course of a VSA computation, as it prevents the error accumulation detrimental in analog computing \citep{marsocci1956error}.

Different VSA models use different types of random vectors. For example, the Binary Spatter Code uses binary vectors \citep{Kanerva1997}, the  Holographic Reduced Representation uses real-valued vectors \citep{plate1995holographic}, and the Frequency domain Holographic Reduced Representation uses complex-valued vectors \citep{plate1995holographic}.   
Most VSA models do not support sparse representations (but see~\citet{Rachkovskij2001}), which are advantageous in terms of energy consumption and usage of synaptic memory. 
Recently, VSA models were proposed that operate with sparse vectors that have block structure \citep{Laiho2015,frady2020variable}, which are related to the Modular Composite Representation VSA model~\citep{SnaiderModular2014}.


The second essential property of VSAs is that all computations can be composed by a small number of elementary vector operations, that, together with the vector representation space, form an algebraic ring structure.  
VSAs feature at least {\it two elementary dyadic operations}, which map two (or more than two, by consecutive execution) vectors into a new vector. 
The {\it bundling operation}:
\begin{equation}
    \mathbf{s} = \mathbf{x}+\mathbf{y}
    \label{bundling}
\end{equation}
is commonly used to represent sets of symbols~\citep{KleykoABF2020}, and is simply the component-wise sum (superposition)\footnote{In some VSA models bundling also involves normalization and/or thresholding of the components of the sum vector, e.g., to keep components of the vector in the binary space~\citep{Kanerva1997} or to preserve the norm of the vector~\citep{plate1995holographic}.} of corresponding symbol vectors. The elements of the set represented by $\mathbf{s}$ can be read out by forming the inner product between the set representation and representations of individual symbols. Depending on the vector dimension, the elements of the set can be detected precisely by a high inner product; for theoretical analysis of this ``readout'', see \citet{Frady2018,thomas2020theoretical, KleykoPerceptron2020}. 

Conversely, the {\it binding operation}:
\begin{equation}
    \mathbf{z} = \mathbf{x} \circ \mathbf{y} 
    \label{binding}
\end{equation}
is commonly used to represent an association between two vectors, such as a key-value pair~\citep{Kanerva2009}. An important property of binding is that there exists the inverse operation, {\it unbinding}, which extracts from the compound data structure $\mathbf{z}$ formed by (\ref{binding}) one of its constituents:
\begin{equation}
     \mathbf{y} \approx \mathbf{x} \unbind \mathbf{z} = \tilde{\mathbf{x}} \circ \mathbf{z}
    \label{unbinding}
\end{equation}
In (\ref{unbinding}), $\tilde{\mathbf{x}}$ denotes the inverse of $\mathbf{x}$ in terms of the binding operation.
The binding operation produces a vector for the association of two or more symbols that, in terms of the inner product, is dissimilar from its arguments. The binding operation $\circ$ has an approximate inverse operation $\unbind$ in the sense that the ``unbound'' vector has the same inner products with other symbol representations as the argument originally used in the binding step.
We will use notation that includes real- and complex-valued vector spaces, i.e., the inner product of two vectors $\mathbf{x}$ and $\mathbf{y}$ will be written as $\mathbf{x}^{\top} \overline{\mathbf{y}}$, where $\overline{\mathbf{y}}$ is the vector with the conjugate complex components of $\mathbf{y}$. 
A general property of VSA models is the following interaction between the binding/unbinding operations and the inner product:
\begin{equation}
    (\mathbf{x} \circ \mathbf{y})^{\top} \overline{\mathbf{z}} = \mathbf{y}^{\top} \overline{(\mathbf{x} \unbind \mathbf{z})} = \mathbf{y}^{\top} \overline{(\tilde{\mathbf{x}} \circ \mathbf{z})},
    \label{bind_innerprod_interaction}
\end{equation}
here $\tilde{\mathbf{x}}$ denotes the inverse of $\mathbf{x}$ with respect to binding.
There are natural combinations of random vector types and specific binding operations, i.e., combinations in which the binding operation preserves the vector type. 
For example, for bipolar \citep{Kanerva2009} or phasor vectors \citep{Frady2018}, the Hadamard product preserves the vector type, whereas for binary vectors \citep{Kanerva1997} it is the component-wise XOR operation. For real- or complex-valued random vectors \citep{plate1995holographic}, circular convolution preserves the vector type, and for block sparse vectors,   
block-local circular convolution \citep{frady2020variable} does so.

The combination of the dyadic vector operations and symbolic representations allows VSA models to represent and query an impressive range of data structures (see~\citet{KleykoComputingParadigm2021} for an overview), such as key-value pairs~\citep{Kanerva2009}, sets~\citep{KleykoABF2020}, histograms~\citep{JoshiNgrams2016}, sequences~\citep{HannaganHolographic2011}, trees~\citep{RachkovskijBinding2001,FradyResonator2020}, stacks~\citep{YerxaUCBHD_FSA2018}, state automata~\citep{OsipovHD_FSA2017}, etc. 

VSAs have been extended to process real-valued data by combining the symbolic and algebraic features with locality preserving encoding (LPE) methods, which have shown promise in applications \citep{PlateRecurrent1992,WeissOlshausenSpatial16, rahimi2017high, FradyDisentangling2018, KomerContinuous2019}, but have previously lacked theoretical underpinning. 
To provide a solid theoretical foundation, we build on fundamental results from 
functional analysis, described in the following section. 


\subsection{Functional analysis results for extending VSA to real-valued data}
\label{funct_ana_background}


In order to process data on a continuous manifold, VSAs have been combined with an LPE method~\citep{PlateRecurrent1992, WeissOlshausenSpatial16}. 
In Section~\ref{symbolic_VSA_kernel}, we show how the encoding in symbolic VSA can be formulated in terms of an inner product similarity kernel. In Section~\ref{sec:lpe_in_vsa}, we generalize the inner product similarity kernel of symbolic VSA to describe an LPE. In this paper, we will focus on kernel LPEs (KLPE), that is, LPEs with a translation-invariant, positive-definite, smoothly decaying similarity kernel (see the definition in Section~\ref{sec:lpe_in_vsa}).
Combining KLPE and VSA produces a computing framework we refer to as Vector Function Architectures (VFA), in which not only symbols but real-valued data and functions can be represented and manipulated in a transparent fashion. The following definitions and results from functional analysis are crucial for understanding the objects that are represented in VFA.  

\vspace{0.3cm}
\noindent
{\it Definition:} A kernel $K(\mathbf{x}_1, \mathbf{x}_2)$ is positive definite if, for any finite set of points $\{\mathbf{x}_1, \mathbf{x}_2, ..., \mathbf{x}_m\}$, the Gram matrix $K(\mathbf{x}_i, \mathbf{x}_j)$ is positive definite (i.e., all eigenvalues are non-negative). 

\vspace{0.3cm}
\noindent
{\it Theorem:} All inner product kernels are positive definite \citep{scholkopf2002learning, hofmann2008kernel}.

\vspace{0.3cm}

Our first central result, Theorem~1, the mathematical interpretation of representations and operations in VFA, will leverage the following theorem by Nachman Aronszajn (1907-1980).

\vspace{0.3cm}
\noindent
{\it Theorem \citep{aronszajn1950theory}:} Each positive definite kernel defines a reproducing kernel Hilbert space (RKHS). 

\vspace{0.3cm}

The following famous theorem of Salomon Bochner (1899-1982) will be used to analyze VFAs and to design models with the desired shapes of similarity kernels.

\vspace{0.3cm}
\noindent
{\it Theorem \citep{bochner1932theory}:} Each continuous kernel $K(\mathbf{x} - \mathbf{y})$ is positive definite if, and only if, it is the Fourier transform of a positive definite measure $p({\boldsymbol \omega})$:
\begin{equation}
    K(\mathbf{x}-\mathbf{y}) = \int d{\boldsymbol \omega}\; p({\boldsymbol \omega}) e^{\imath \;{\boldsymbol \omega}^{\top} (\mathbf{x}-\mathbf{y})} = E_{p({\boldsymbol \omega})}\left[ 
    e^{\imath \;{\boldsymbol \omega}^{\top} \mathbf{x}}
    \overline{e^{\imath \;{\boldsymbol \omega}^{\top} \mathbf{y}}}
    \right]
\label{bochner}
\end{equation}
Proof: For formulation of this theorem and proof, see~\citet{rudin1962fourier}.  
\vspace{0.3cm}

Kernel methods have been first used in nonlinear pattern recognition by \citet{aizerman1964theoretical} and became a corner stone of machine learning in the 1990ies.
For an excellent textbook on functional analysis of kernel methods and its applications in machine learning, see \citet{scholkopf2002learning}.




\section{Vector Function Architecture (VFA)}
\label{sec:VFA}

In this section, we define and characterize a generalization of VSA for computing with functions, which shares its transparency with the original symbolic VSA described in Section~\ref{sec:background:symbolic:VSA}. Our generalization of VSA starts from reformulating symbolic VSA in terms of an inner-product similarity kernel described in the following subsection. 

\subsection{The kernel of symbolic VSAs}
\label{symbolic_VSA_kernel}
For symbolic operations (e.g., for analogical reasoning~\citep{PlateAnalogical1994,RachkovskijAnalogy2012}) the encoding scheme $s \to \mathbf{z}(s) \in \mathbb{C}^n$ should guarantee the best possible separation of different symbols by their high-dimensional projections, i.e., by forming inner products between representations $\mathbf{z}(s)$. Further, the binding operation associating vectors of symbols $s_1$ and $s_2$ creates a new representation vector $\mathbf{z}(s_1) \circ \mathbf{z}(s_2)$, which is well separated from both of the representations $\mathbf{z}(s_1)$ and $\mathbf{z}(s_2)$. This property is crucial because it enables the construction of compound data structures ``on the fly''.

An optimal separation can be achieved by an encoding scheme with exactly orthogonal representation vectors. The inner product of orthogonal vectors constitutes, up to a normalization factor, the {\it ideal similarity kernel} for symbolic computation: 
\begin{equation}
    \mathbf{z}(s_1)^{\top} \overline{\mathbf{z}(s_2)} \overset{\mbox{\tiny large } n}{\longrightarrow} K_{Kron}(s_1,s_2) := \delta_{s_1, s_2},
\end{equation}
with $\delta_{u,v}$ being the Kronecker delta function. This encoding scheme, however, is limited to maximally $n$ symbols, a quite inefficient use of an $n$-dimensional representation space. 

Therefore, symbolic VSA models typically employ encoding schemes with pseudo-orthogonal random vectors, sampled from some distribution $p(\mathbf{z})$. 
This encoding scheme allows expression of many more symbols than dimensions in the vector space.
The inner product still equals the ideal kernel value for $s_1=s_2$, however, for $s_1 \not = s_2$ it only approximates its value of $0$: 
\begin{equation}
    \left| \mathbf{z}(s_1)^{\top} \overline{\mathbf{z}(s_2)} - K_{Kron}(s_1,s_2)\right|
    \leq \epsilon(n)
\end{equation}
with $\epsilon(n) \to 0$ for increasing $n$.
The convergence of pseudo-orthogonality to exact orthogonality, i.e., of the inner product to the ideal Kronecker delta kernel, is referred to in statistics as {\it concentration of measure}  \citep{ledoux2001concentration}, 
and can be analyzed rigorously \citep{Frady2018,thomas2020theoretical}. 

The correction of these errors, as well as noise from other sources, is an important part of VSA computation -- it prevents uncontrollable error accumulation, a major problem in analog computers \citep{marsocci1956error}.
For error correction in VSA, a vector, occurring in the input or in the course of algebraic reasoning, is compared with the base vectors assigned to symbols. 
In symbolic VSA, detection and denoising is typically accomplished by a content-addressable memory \citep{plate1995holographic,GritsenkoAMSurvey2017}. The correctness of the detection process depends on the amount of noise and can be predicted by the signal detection theory \citep{Frady2018}. 



\subsection{Locality preserving encoding with a kernel}
\label{sec:lpe_in_vsa}


Typical VSA representations with independent random vectors encode via the inner product only binary information, i.e., ``same'' versus ``different'', about the relationship between the encoded objects.   
Encoding graded similarity information between objects requires a dedicated method of locality preserving encoding (LPE). An LPE produces vector representations of points on a manifold, so that the inner product of the vectors reflects 
the relationship between the points. 
This allows data to be expressed and manipulated in the context of a data manifold. 

Here we focus on the combination of VSA with a particular type of LPE that induces by the inner product of the encoding vectors a translation-invariant similarity kernel. 
Generalizing our treatment of encoding functions for symbolic VSA in Section~\ref{symbolic_VSA_kernel}, we can define the following type of LPE functions: 

\vspace{0.3cm}
\noindent
{\it Definition 1:} A randomizing LPE encoding function $f: r\in \mathbb{R} \to \mathbf{z}(r) \in \mathbb{C}^n$ is a {\it kernel-LPE (KLPE)}, that is, it induces a similarity kernel, if the following requirement holds:

\begin{itemize}

\item[] 

In the limit for large dimensionality $n$, the {\it inner product} between point representations defines a translation-invariant similarity kernel:
    \begin{equation}
        \mathbf{z}(r_1)^{\top} \overline{\mathbf{z}(r_2)}  
        \overset{\mbox{\tiny large } n}{\longrightarrow} K(r_1-r_2),
        \label{kernelconvergence}
    \end{equation}
with a kernel function $K(d)$, which is real-valued, assumes its maximum at $d=0$ and gradually reaches zero for large $|d|$.
Under quite general conditions, the convergence in (\ref{kernelconvergence}), is fast:
\begin{equation}
    \mbox{Pr}\left(\left| \mathbf{z}(r_1)^{\top}\overline{\mathbf{z}(r_2)} - K(r_1-r_2)\right| > \epsilon\right) <  g(m, \epsilon),
    \label{kernelconv2}
\end{equation}
with $g(m, \epsilon)$ a positive function that for any fixed $\epsilon$ approaches zero exponentially fast with growing $n$~\citep{rahimi2007random}. The empirical convergence of kernels with increasing vector dimension is shown in the Figures~3, 5 and 6 below. For specific algorithms we will discuss the dependency between vector dimension and computing precision in Section~\ref{sec:vfa_nonpara_ker_met}.

\end{itemize}

Representations of data points formed by a KLPE have the following properties. For a large distance between the data points $r_1$ and $r_2$, the resulting representations are pseudo-orthogonal, just as in symbolic VSA. Conversely, for a small distance, the corresponding representation vectors have a systematic correlation.

\subsection{Definition and properties of VFA}

Based on the theorem of \citet{aronszajn1950theory}, stated in Section~\ref{funct_ana_background}, we can concisely define the combination of VSA and KLPE and characterize its properties. We start with the following definitions:

\vspace{0.3cm}
\noindent
{\it Definition 2:} A KLPE {\it is compatible with a VSA binding operation} $\circ$, if the addition of two values of the encoded variable can be represented by binding the individual representations of the values:
\begin{equation}
    \mathbf{z}(r_1+r_2) =  \mathbf{z}(r_1) \circ \mathbf{z}(r_2). 
    \label{transbybind}
\end{equation}

\vspace{0.3cm}
\noindent
{\it Definition 3:} The combination of a VSA with a KLPE that is compatible with the VSA binding operation induces an RKHS of functions with inner product $\left< f, g \right>:=\int_{-\infty}^{\infty} f(r) g(r) dr$ that we will call a {\it Vector Function Architecture (VFA)}: 
\begin{itemize}
    \item[a)] 
A function of the form: 
\begin{equation}
    f(r) = \sum_k \alpha_k K(r-r_k)
    \label{expand}
\end{equation}
is represented via (\ref{kernelconv2}), $f(r) = \mathbf{y}_{f}^\top \overline{\mathbf{z}(r)}$, by the vector:
\begin{eqnarray}
     \mathbf{y}_f &=& \sum_k \alpha_k \mathbf{z}(r_k)
     \label{eq:vfa:function}
\end{eqnarray}

\item[b)] The reproducing kernel property is $[f*K](u) = \left< f, K_u \right> = f(x+u)$, with $K_u := K(x-u)$ and in particular $\left<K_s, K_u \right> = [K \ast K](s-u) = K(s-u)$ \citep{hofmann2008kernel}.

\end{itemize}


We can now ask how different vector operations act on functions represented in the VFA. The following theorem summarizes the function operations available in VFA. 

\vspace{0.3cm}
\noindent
{\it Theorem 1:}  In VFA, the following operations in the corresponding RKHS function space can be computed by elementary VSA vector operations:
\begin{itemize}
\item {Point-wise readout of a function} is the inner product between the VSA representation of the function and the LPE representation of point $s$:
\begin{equation}
    f(s) = \left<f,K_s\right> = \mathbf{y}_{f}^\top \overline{\mathbf{z}(s)}
    \label{point_readout}
\end{equation}
with $K_s := K(r-s)$.
\item {Point-wise addition of functions} is done by the bundling operation (\ref{bundling}):  
\begin{equation}
    \mathbf{y}_{f+g} = \mathbf{y}_f + \mathbf{y}_g
    \label{functadd}
\end{equation}

\item {Function shifting} $f(x) \to g(x) = f(x+r)$ is done by VSA binding (\ref{binding}) between a function vector and the point representation of the translation vector:
\begin{equation}
    \mathbf{y}_{g} = \mathbf{y}_f \circ \mathbf{z}(r)
    \label{functshift}
\end{equation}

\item {Function convolution} is done by the VSA binding operation:
\begin{equation}
    \mathbf{y}_{f*g} = \mathbf{y}_f \circ \mathbf{y}_g
    \label{functconv}
\end{equation}

\item {Overall similarity between functions}, the inner product in function space, is equal to the inner product in VFA space:
\begin{equation}
    \left<f,g\right> =  \mathbf{y}_f^{\top} \overline{\mathbf{y}_g}
\label{functinprod}
\end{equation}
\end{itemize}

\vspace{0.3cm}
\noindent
{\it Proof:}\\ 

\noindent
{Point-wise readout of a function} (\ref{point_readout}) is part of the Definition of VFA.\\

\noindent
{Point-wise addition of functions} (\ref{functadd}) follows from linearity of inner product kernels.\\

\noindent
{Function shifting} (\ref{functshift}) follows from the RKHS properties, see Definition VFA, and the compatibility of the KLPE with binding (\ref{transbybind}).\\ 

\noindent
{Function convolution} (\ref{functconv}) can be
is computed as: 
\begin{eqnarray}
    [f*g](u) &=& \int_{-\infty}^{\infty} f(r) g(u-r) dr = \left< f, g_u \right> = \sum_{k,l} \alpha_k \beta_l K(r_k+r_l+u)\nonumber\\
    &=& \sum_{k,l} \alpha_k \beta_l \mathbf{z}(r_k)^{\top} \overline{\mathbf{z}(-r_l+u)} = \left(\sum_{k} \alpha_k \mathbf{z}(r_k)\right)^{\top} \overline{\sum_{l} \beta_l \mathbf{z}(-r_l+u)}\nonumber\\
    &=& \left(\sum_{k} \alpha_k \mathbf{z}(r_k)\right)^{\top} \overline{\sum_{l} \beta_l \mathbf{z}(-r_l)\circ \mathbf{z}(u)}
    = \left(\sum_{k} \alpha_k \mathbf{z}(r_k)\right)^{\top} \overline{\sum_{l} \beta_l \tilde{\mathbf{z}}(r_l)\circ \mathbf{z}(u)}\nonumber\\
    &=& \mathbf{y}_f^{\top} \overline{\tilde{\mathbf{y}_g}\circ \mathbf{z}(u)}
    = \left[\tilde{\tilde{\mathbf{y}_g}} \circ \mathbf{y}_f\right]^{\top} \overline{\mathbf{z}(u)}
    = \left[\mathbf{y}_g \circ \mathbf{y}_f\right]^{\top} \overline{\mathbf{z}(u)}
    \label{convformula}
\end{eqnarray}
In (\ref{convformula}) we use $\tilde{\mathbf{y}}$ to denote the inverse of $\mathbf{y}$ with respect to the binding operation.
The transformations here used (\ref{transbybind}) and (\ref{bind_innerprod_interaction}).\\

\noindent
{Overall similarity between functions} (\ref{functinprod}) can be calculated as:
\begin{equation}
    \left<f,g\right> = \int_{-\infty}^{\infty} f(r) g(r) dr = \sum_{k,l} \alpha_k \beta_l K(r_k-r_l) 
    = \sum_{k} \alpha_k \mathbf{z}(r_k) \sum_{l}\beta_l \overline{\mathbf{z}(r_l)}
    = \mathbf{y}_f^{\top} \overline{\mathbf{y}_g}
\end{equation}\\


\noindent
$\Box$\\ 

As we will discuss later on, VFAs can equally be defined on multi-dimensional spaces, just by using a KLPE method for multi-dimensional input. The corresponding kernel function is:
\begin{equation}
    K(\mathbf{r}-\mathbf{s}) = \mathbf{z}(\mathbf{r})^{\top} \overline{\mathbf{z}(\mathbf{s})}
\end{equation}
and the represented functions are of the form $f(\mathbf{s}) = \sum_k \alpha_i K(\mathbf{r}^k-\mathbf{s})$. The same function operations as described above can be performed in such VFAs for functions with a multi-dimensional domain.

\section{VFA with fractional power encoding (FPE)} 
\label{sec:VFA:FPE}

The previous section defined and described VFA in general. A prerequisite of a VFA is a KLPE which is compatible with a VSA binding operation. 
We now focus on a particular LPE, fractional power encoding (FPE), which is based on a binding operation, and was already introduced in the early VSA literature~\citep{Plate1994}.  We describe the different flavors of FPE, show that they are KLPEs compatible with binding. These findings suggest that earlier models combining VSA with FPE do qualify as VFAs - although their ability to represent and manipulate functions was not exploited exlicitly. 


\subsection{Generalization of Plate's fractional power vector}
\label{sec:VSA:LPE:frac}

FPE is a generalization of the fractional power vector \citep{PlateRecurrent1992, Plate1994}, an LPE method for encoding topological spaces based on circular convolution binding. Here we generalize the concept to other VSA binding operations. FPE starts from self-binding, i.e., binding a base vector, a random vector $\mathbf{z} \sim p(\mathbf{z})$, $i$ times with itself, which defines an encoding strategy for integers:
\begin{equation}
    \mathbf{z}(i) := (\mathbf{z})^{(\circ i)} =
    \circ_{j=1}^i  \mathbf{z}=
    \mathbf{z} \circ ... \circ \mathbf{z}
    \label{int_exp_binding}
\end{equation}
Here $\circ$ is a generic VSA binding operation. Note that by the properties of VSA binding, self-binding creates pseudo-orthogonal representation vectors $\mathbf{z}(i)$ for different integers $i$. 

\begin{figure}[H]
\begin{center}
 \begin{tikzpicture}[fill=gray]
\scope
\clip  
(1,0) circle (1);
\fill[orange] (0,0) circle (1);
\endscope
\scope
\clip (-2,-2) rectangle (2,2)
      (0,0) circle (1);
\fill[green] (1,0) circle (1);
\endscope
\draw [black,dashed] (0,0) circle (1) (0,1)  node [text=black,below] {$A_{\circ}$};
\draw (1,0) circle (1) (1.5,0.5)  node [text=black,below] {$\mathbf{Y}_{\mathbf{z}}$}
       (0.5,0.28) ellipse (90pt and 45pt)
(0.5,1.6) node [text=black,below] {Entire VFA vector space}
(0.5,0.5) node [text=black,below] {$\mathbf{P}_{\mathbf{z}}$};
\draw (0.5,-0.2) node [text=black,below] {$\times \;\mathbf{z}$};
\end{tikzpicture}
\vspace{-1cm}
\caption{Visualization of the representation space in a VFA based on FPE. The elipse symbolizes the entire space of representation vectors. The dashed circle symbolizes $A_{\circ}$, the set of unitary vectors with respect to a specific binding operation $\circ$. The point $\mathbf{z}$, marked by $\times$, is the random base vector of a specific FPE $\mathbf{z}(r)$. The solid circle symbolizes the subset of vectors that can be interpreted in the context of this specific FPE: The subset $\mathbf{P}_{\mathbf{z}} := \{\mathbf{z}(r): \forall r\}$ contains the representations of single points that are encoded by unitary vectors, lying  in $A_{\circ}$. The set $\mathbf{Y}_{\mathbf{z}}$ contains the linear superpositions of elements in $\mathbf{P}_{\mathbf{z}}$, these are non-unitary vectors that represent other elements of the function space associated with the specific VFA.}
\label{fig:fvsa_visu}
\end{center}
\end{figure}
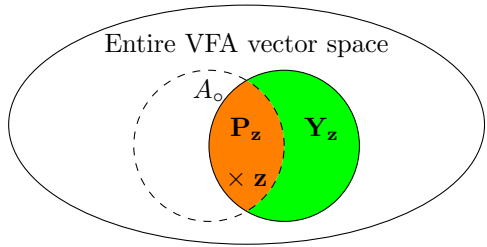

FPE \citep{PlateRecurrent1992} is the generalization of self-binding (\ref{int_exp_binding}) from integer exponents to real-valued exponents  (see also Section 5.6 in~\citet{Plate1994}):
\begin{equation}
    f_{\mathrm{FPE}}: r\in \mathbb{R} \to \mathbf{z}(r) = (\mathbf{z})^{(\circ r)} \in \mathbb{C}^n
    \label{fpe-code}
\end{equation}
Let us denote the set of norm-preserving vectors with respect to the binding operation  $\circ$ with:
\begin{equation}
    A_{\circ} := \{\mathbf{z}: ||\mathbf{v} \circ \mathbf{z}||^2 = ||\mathbf{v}||^2 \;\forall \;\mathbf{v}\}
\end{equation}
Note that with the base vector chosen in $A_{\circ}$, all points $r$ are mapped by (\ref{fpe-code}) to vectors also within $A_{\circ}$. Further, if the base vector is sampled from the probability distribution $p(\mathbf{z})$, representations for values other than $r=1$ are random vectors described by other distributions than the base vector. In particular, the representation $\mathbf{z}(0)=(\mathbf{z})^{(\circ 0)}$ is the identity vector with respect to the binding operation, the same deterministic vector for any base vector.  
The full vector space of a VFA can be visualized (Figure~\ref{fig:fvsa_visu}). 
The dashed circle marks the subset $A_{\circ}$ of unitary vectors. For one specific FPE, a fixed base vector $\mathbf{z}$, marked by $\times$, is sampled from a distribution $p(\mathbf{z})$. A subset of vectors, symbolized by the solid circle, can be decoded or interpreted in the context of this FPE. The part of this set that overlaps with $A_{\circ}$, $\mathbf{P}_{\mathbf{z}} := \{\mathbf{z}(r): \forall r\}$, contains the representations of single points. The other part of the set, $\mathbf{Y}_{\mathbf{z}}$, contains non-unitary vectors that represent functions. Vectors outside the solid circle cannot be interpreted in the context of the one specific FPE, they could be used to encode other objects, such as symbols or function spaces defined by other FPEs.
Thus, for a given arbitrary vector occurring in the input or during a calculation, there is a detection and decoding problem in VFA, which is similar to the detection problem in symbolic VSA. We will propose a solution to this in Section~\ref{sec:VFA:detection}.

\subsection{Existing binding operations induce different types of VFA}

FPE hinges critically on the VSA binding operation and the properties of the vector sets depicted in Figure~\ref{fig:fvsa_visu} depend on the individual binding operation. In the following we define and analyze the properties of FPEs based on previously proposed binding operations, the Hadamard product, circular convolution or block-local circular convolution. For the corresponding VFAs, the definitions yield concrete formulae for computing the function manipulations in Theorem~1, as well as for the sets depicted in Figure~\ref{fig:fvsa_visu}.

\subsubsection{VFA with Hadamard product}

The Hadamard product between two vectors is the vector of the multiplied individual  vector components:
\begin{equation}
    \mathbf{z} = \mathbf{x} \odot \mathbf{y} := (x_1 y_1,....x_n y_n)^{\top} \;\;\; ;\;\;\; \mathbf{y} = \mathbf{x} \hadarunbind \mathbf{z} = \tilde{\mathbf{x}} \odot \mathbf{z} = \overline{\mathbf{x}} \odot \mathbf{z}
\end{equation}
An {\it FPE based on Hadamard product} encodes a value $r \in \mathbb{R}$ by:
\begin{equation}
    \mathbf{z}^{hp}(r) :=  (\mathbf{z})^r 
    \label{frac_exp_binding}
\end{equation}
where $\mathbf{z} \in A_{\odot}$ is a fixed base vector, and $(\mathbf{z})^r$ denotes the exponentiation of each vector component by $r$. The set of unitary vectors with respect to Hadamard binding is:
\begin{equation}
    A_{\odot} := \{\mathbf{z}: ||\mathbf{v} \odot \mathbf{z}||^2 = ||\mathbf{v}||^2 \;\forall \;\mathbf{v}\} = \{ \mathbf{z}: |z_i| = 1 \;\forall \; i\},
\end{equation}
the set of all phasor vectors, $z_i = e^{\phi_i}$. Commonly, a base vector is determined by sampling each component independently from a uniform phase distribution:
\begin{equation}
  p(\phi_i) = 
    \begin{cases}
      \frac{1}{2\pi} & \text{if $\phi_i \in \{-\pi, \pi\}$}\\
      0 & \text{else}
    \end{cases}       
    \label{uniformphasedistrib}
\end{equation}
but see Section~\ref{sec:shapingFPEkernels} below for other choices of $p(\mathbf{z})$.

Let us briefly return to symbolic processing in classical Hadamard VSAs \citep{Gayler1998, Kanerva2009}. In these models, the components of symbol representations are drawn i.i.d. from a probability distribution, i.e., $p(\mathbf{z}) = \prod_i p(z_i)$.
For encoding two symbols $s_1$ and $s_2$ with vectors $\mathbf{z}$ and $\mathbf{y}$, the joint probability for each pair of components  $z_i, y_i$ is:
\begin{equation}
    p(z, y|s_1, s_2) =
    \begin{cases}
      p(z|s_1)\;\delta(z-y) & \text{if $s_1=s_2$}\\
      p(z|s_1) p(y|s_2) & \text{otherwise}
    \end{cases}     
\label{comp_prob}
\end{equation}
Thus, for a large enough vector dimension $n$, the inner product converges to the expectation defined by the probability distribution (\ref{comp_prob}): 
\begin{equation}
    \frac{1}{n}\mathbf{z}(s_1)^{\top} \overline{\mathbf{z}(s_2)} 
    \overset{\mbox{\tiny large } n}{\longrightarrow}
    E_{p(z, y|s_1, s_2)}[z\; \overline{y}] =
    \begin{cases}
      E_{p(z|s_1)}[z\; \overline{z}] & \text{if $s_1=s_2$}\\
      E_{p(z|s_1) p(y|s_2)}[z\; \overline{y}] & \text{if $s_1 \not=s_2$}
    \end{cases}   
    \label{eq:inner:product}
\end{equation}
Similarly, in the Hadamard FPE (\ref{frac_exp_binding}), vector components $\mathbf{z}(r)$ can be described by a probability distribution. But in this case the distribution depends on the distribution from which the base vector was sampled (\ref{uniformphasedistrib}), as well as on the particular encoded value: $\forall i: z_i \sim p(z_i |r)$.  For a large enough vector dimension $n$, this distribution again determines exactly the inner product of two vectors:
\begin{equation}
    \frac{1}{n}\mathbf{z}(r_1)^{\top} \overline{\mathbf{z}(r_2)} 
    = E_{p(z| r_1) p(y| r_2)}\left[z \overline{y}\right].
    \label{val_sim}
\end{equation}
This relation will be used below to determine the kernel shape of Hadamard FPEs. 

Hadamard FPEs require a complex state space. An often used VSA framework \citep{Gayler1998, Kanerva2009} uses  real-valued components with Hadamard product binding where the norm-preserving vectors are bipolar vectors. 
However, one cannot define an FPE forming real-valued vectors based on the Hadamard product. 
First, all even and all odd integer powers of $r$ are mapped to the same representation vector. Second, non-integer values of $r$ produce complex vectors.
Note that bipolar vectors are a special case of complex vectors with a particular sampling distribution; we will return to discuss how different distributions affect the properties of VFAs.

\subsubsection{VFA with circular convolution}
\label{sec:VFAcc}
Circular convolution is the standard convolution operation used in the discrete finite Fourier transform ${\cal F}$ which can be used to produce a vector from two input vectors $\mathbf{x}$ and $\mathbf{y}$:
\noindent
\begin{eqnarray}
    (\mathbf{x} \circledast \mathbf{y})_k &:=& \sum_{i=1}^N x_{(i-k)_{\mbox{\tiny mod} N}} y_{i} 
    = \mathbf{1}^{\top} \left[\mathbf{x}\mathbf{y}^{\top} \odot \mathbf{P}_k\right] \mathbf{1}
    = \mbox{Tr}\left(\left[\mathbf{x}\mathbf{y}^{\top} \odot \mathbf{P}_k\right] \mathbf{P}_{-k}\right)
    = \left({\cal F}^{-1}\left( \mathbf{X}\odot\mathbf{Y} \right)\right)_k
    \label{circonv}\\
    (\mathbf{x} \ccunbind \mathbf{y})_k &:=& \left({\cal F}^{-1}\left( \overline{\mathbf{X}}\odot\mathbf{Y} \right)\right)_k\nonumber
\end{eqnarray}
where, $\mathbf{P}_k$ is the cyclic index shift permutation by $k$, i.e., $\mathbf{P}_0=\mathds{1}$, $\mathbf{1}$ is the all one vector. The bold capital letters denote the discrete Fourier transform of the corresponding bold lower-case letters, e.g., ${\cal F}(\mathbf{x})=\mathbf{X}$. The first proposal of FPE was based on circular convolution and referred to as the fractional power vector \citep{PlateRecurrent1992,Plate1994}. {\it FPE based on circular convolution} is defined:
\begin{eqnarray}
    \mathbf{z}^{cc}(r) := \mathbf{z}^{(\circledast r)} 
    &=& {\cal F}^{-1}\left( {\cal F}(\mathbf{z})^r \right)
    = F^{-1} \left(F \mathbf{z} \right)^r 
    \label{circonv_fract_binding}
\end{eqnarray}
The set of unitary vectors with respect to circular convolution are vectors whose Fourier transforms are phasor vectors $A_{\circledast} := \{\mathbf{z}: |(F \mathbf{z})_i|=1 \;\forall i \}$:
\begin{equation}
    ||\mathbf{v} \circledast \mathbf{z}||^2 
    = ||F^{-1} \left(F \mathbf{v} \cdot F \mathbf{z} \right) ||^2
    = ||\left(F \mathbf{v} \cdot F\mathbf{z} \right)||^2\nonumber
    \overset{\mathbf{z}\in A_{\circledast}} {=} ||\mathbf{v}||^2 
\end{equation}
Here $F$ and $F^{-1}$ are the Vandermonde matrices representing the discrete Fourier transform and its inverse. 

The unitary vectors in $A_{\circledast}$ are of the form:
\begin{equation}
      F^{-1} (e^{\imath \phi_0}, e^{\imath \phi_1},..., e^{\imath \phi_{n-1}})^{\top} = \frac{1}{\sqrt{n}} \left( \sum_k e^{\imath (2\pi 0 k + \phi_k)}, \sum_k e^{\imath (2\pi 1 k + \phi_k)},..., \sum_k e^{\imath (2\pi (n-1) k + \phi_k)}\right)^{\top} 
\label{circ_unitary_vecs}
\end{equation}
for an arbitrary phasor vector in the Fourier domain with phase angles $\phi_0,..., \phi_{n-1}$. 
If the phasor in Fourier space is sampled from a flat phase distribution (\ref{uniformphasedistrib}), the inverse Fourier transform results in complex-valued vector components where real and imaginary parts appear to be distributed like Gaussian random variables. However, producing base vectors by independently sampling each component from a Gaussian distribution, as originally described by \citet{Plate1991}, is not a good strategy because these vectors are unlikely to be exactly unitary. 

A notable subset of the unitary vectors $A_{\circledast}$ are one-hot phasor vectors $(0,0,...,z_l, ... 0)^{\top}$, which are unitary because their Fourier transforms are dense phasor vectors: 
\begin{equation}
    F \mathbf{z} = z_l (e^{-\frac{2\pi\imath}{n} 0 l}, ..., e^{-\frac{2\pi\imath}{n} (n-1) l})^{\top} = z_l F_l^{\top}
    \label{ohotFT}
\end{equation}
where $F_l^{\top}$ is the $l$-th column of the discrete Fourier matrix. Thus, when the Fourier vectors are columns of the discrete Fourier matrix (with constant phase offset) the corresponding base vectors are one-hot. 

Further, it is often desired that the base vectors used in circular convolution are purely real-valued. This requires Hermitian symmetry of the random vectors sampled in the Fourier domain, that is, a phasor entry for a negative frequency must be the complex conjugate of the phasor entry for the corresponding positive frequency \citep{KomerContinuous2019}.

\subsubsection{VFA with block-local circular convolution}
\label{sec:blolocico}
The FPE base vectors of VFAs described so far are either fully dense phasors in models with Hadamard product binding, or they can be maximally sparse, i.e., one-hot phasors, in models with circular convolution binding. One can build VFAs on a middle ground, in which the FPE base phasor vector has a desired sparsity level of $k<n$. In this construction, the base vectors are {\it sparse block vectors}, with $k$ equally sized compartments, in which each compartment is a one-hot phasor. Previously, VSA with such sparse block vectors have been proposed that use so-called block-local circular convolution (LCC) as the binding operation \citep{frady2020variable}:
\begin{eqnarray}
    (\mathbf{z}\ast_{\!B} \mathbf{y})_{(\mbox{\tiny block } i)} &:=& \mathbf{z}_{(\mbox{\tiny block } i)} \circledast \mathbf{y}_{(\mbox{\tiny block } i)}
    = F^{-1} \left(F\mathbf{z}_{(\mbox{\tiny block } i)} \odot F\mathbf{y}_{(\mbox{\tiny block } i)}\right)
    \label{lcc_bind}\\
    (\mathbf{z} \lccunbind \mathbf{y})_{(\mbox{\tiny block } i)} &:=& F^{-1} \left(\overline{F\mathbf{z}_{(\mbox{\tiny block } i)}} \odot F\mathbf{y}_{(\mbox{\tiny block } i)}\right)\nonumber
\end{eqnarray}
where $F$ is the Vandermonde matrix representing the discrete Fourier transform for a block of size $n/k$. In essence, the binding operation (\ref{lcc_bind}) forms a new one-hot vector for each block, the index of the active element is just the sum of indices of the hot components of the input vectors modulo block size~\citep{Laiho2015, frady2020variable}.  

\begin{figure}[H]
    \centering
    \includegraphics[width=0.6\textwidth]{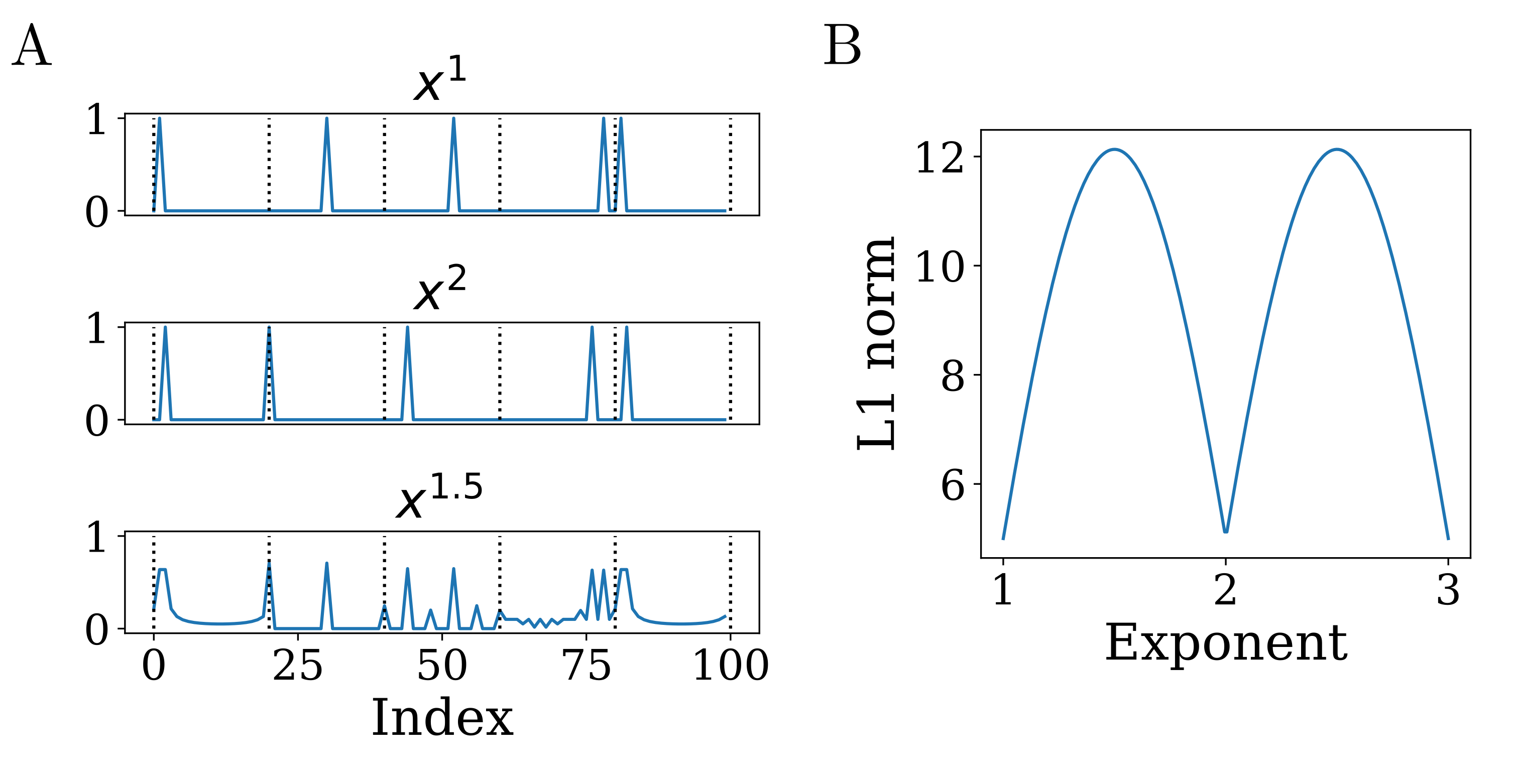}
    \caption{A. Representations of different values ($r= 1, 2, 1.5$) produced by an LCC FPE with a $k$-sparse ($k=5$) base vector (upper panel; plots show magnitude). The representations of integer values  are exactly $k$-sparse (upper and middle panel), the representations of non-integer values are less sparse (lower panel). B. $L_1$ norm of vectors produced by LCC FPE as a function of encoded values between 1 and 3.}
    \label{fig:sparse_fpe}
\end{figure}

{\it FPE based on LCC} is defined by generalizing self-binding with LCC:
\begin{equation}
    \mathbf{z}^{lcc}(r)_{(\mbox{\tiny block } i)} := \mathbf{z}^{(\ast_{\!B}\, r)}_{(\mbox{\tiny block } i)} 
    = F^{-1} \left(F \mathbf{z}_{(\mbox{\tiny block } i)} \right)^r 
    \label{local_circonv_fract_binding}
\end{equation}
The properties of the set of unitary vectors $A_{\ast_{\!B}}$ can be understood by realizing that each block of a unitary vector has to be unitary with respect to circular convolution, see Section~\ref{sec:VFAcc}. Thus, the unitary vectors include $k$-sparse phasor vectors, in which each block is exactly one-hot, but also vectors that are denser than $k$-sparse vectors. 
For producing as sparse as possible representation vectors we sample the base vector of an LCC FPE from a distribution $p(\mathbf{z})$ that produces an exactly $k$-sparse random vector in $A_{\ast_{\!B}}$.
The representations of all integer values of $r$ are then again exactly $k$-sparse, for intermediate values, the sparsity is slightly lower as shown in Figure~\ref{fig:sparse_fpe}.

\section{The kernels of FPE}
\label{sec:FPE:kernels}

We have seen in Section~\ref{sec:VFA:FPE} that the combination of a VSA and an FPE induces a VFA. Although such models have been described earlier, their power in representing certain classes of functions and manipulating them has not been exploited explicitly. In this section, we analyze the kernel properties that can be realized in such VFA models. 

\subsection{FPEs with uniformly sampled base vectors have a universal kernel}
Formally, FPE looks quite different for the different realizations of the binding operation. Depending on the realization it also requires different base vectors. Nevertheless, FPE possesses a universal similarity kernel, as stated in the following theorem.   

\vspace{0.5cm}
\noindent
{\it Theorem 2:} Assume an FPE with a uniformly sampled base vector, which is the typical procedure for sampling VSA vectors. For a Hadamard FPE this means the phases of the base vector are sampled from the uniform phase distribution, whereas for CC FPE and LCC FPE this means the base vector corresponds in the Fourier domain to a random phasor vector with samples from the uniform phase distribution. The FPE then induces a VFA which is the RKHS of band-limited continuous functions, independent of the underlying realization of the binding operation. Specifically, the kernel of FPE is the sinc function, which defines the RKHS of the band-limited continuous functions.    

\vspace{0.5cm}
\noindent
{\it Proof:}
\begin{itemize}
    \item[1)] {\it Hadamard product:} Using (\ref{val_sim}), the inner product of the representations of $r$ and $r+d$ can be written: 
\begin{align}
\begin{split}
    \frac{1}{n}(\mathbf{z}(r+d))^{\top} \overline{\mathbf{z}(r)} 
    &= \frac{1}{n}(\mathbf{z}^{r+d})^{\top} \overline{\mathbf{z}^{r}} 
    = \frac{1}{n}(\mathbf{z}^{d})^{\top}\mathbf{1}
    = \frac{1}{n}\sum_j e^{\imath \phi_j d} 
    \to E_{p(\phi)}[e^{\imath \phi d}] \\
     &= 2 \pi \int_{\omega} p(2 \pi \omega) e^{\imath 2 \pi \omega d} d\omega 
    = 2 \pi {\cal F}[p(2 \pi \omega)](d)  
    \label{kerninner}
\end{split}
\end{align}
where $\mathbf{1}$ is the vector of $1$-entries. Note that the inner product (\ref{kerninner}) is the average over $n$ random phasors $e^{\imath \phi d}$. For large $n$, this average can be rewritten as the expectation of these random phasors under the uniform phase distribution (\ref{uniformphasedistrib}) used to generate the phases $\phi_j$ of the base vector $\mathbf{z}$.
By renaming the phase variable $\phi$ as a frequency $2 \pi \omega$, the expectation becomes a Fourier integral. Expressing the kernel as the Fourier transform of a distribution is an application of Bochner's theorem (\ref{bochner}).
Thus, the inner product kernel (\ref{kerninner}) converges to the Fourier transform of a rectangular frequency distribution $2 \pi p(2 \pi \omega)$ with $p(\phi)$ from (\ref{uniformphasedistrib}). The result is the sinc function:
\begin{equation}
    K_{\odot}(d) \to {\cal F}[p(\omega)](d) = \frac{1}{\sqrt{2\pi}}\mbox{sinc}(\pi d)
    \label{fhp_kernel}
\end{equation}
This result is well known in statistical optics, as the averaging of phasors is a problem often encountered in optics \citep{goodman2007speckle, goodman2015statistical}.

\item[2)] {\it Circular convolution:}
The inner product of the representations of $r$ and $r+d$ can be written: 
\begin{equation}
    (\mathbf{z}^{(\circledast (r+d))})^{\top} \overline{\mathbf{z}^{(\circledast r)}}
    =  \left( F^{-1} (F \mathbf{z})^{(r+d)}\right)^{\top} 
    \overline{F^{-1} \left(F \mathbf{z} \right)^{r} }
    =  \left((F \mathbf{z})^{(r+d)}\right)^{\top} 
    \overline{\left(F \mathbf{z} \right)^{r} }
    =\sum_j (F \mathbf{z})_j^{d}  
    \label{cckerninner}
\end{equation}
Thus, the inner product is again a sum of $n$ random phasors $e^{\imath \phi d}$
with $\phi \sim p(\phi)$ from (\ref{uniformphasedistrib}),
and the inner product kernel $K_{\circledast}(d)$ for circular convolution binding converges for large $n$ to the same kernel as for FPE with Hadamard product (\ref{fhp_kernel}), the sinc function, see also~\citet{voelker2020short}.

\item[3)]{\it Block-local circular convolution:}
The inner product of the representations of $r$ and $r+d$ can be written: 
\begin{equation}
    \frac{1}{k} \sum_{i=1}^{k} (\mathbf{z}^{(\ast_{\!B}\, (r+d))}_{(\mbox{\tiny block } i)})^{\top} \overline{\mathbf{z}^{(\ast_{\!B}\, r)}_{(\mbox{\tiny block } i)}}
    = \frac{1}{k} \sum_{i=1}^{k} \left((F \mathbf{z}_{(\mbox{\tiny block } i)})^{(r+d)}\right)^{\top} 
    \overline{\left(F \mathbf{z}_{(\mbox{\tiny block } i)} \right)^{r} }
    =\frac{1}{k}\sum_{i=1}^{k} \sum_{j=1}^{n/k} (F \mathbf{z}_{(\mbox{\tiny block } i)})_j^{d}  
    \label{bskerninner}
\end{equation}
Again, the inner product is a sum of $n$ random phasors $e^{\imath \phi d}$ 
with $\phi \sim p(\phi)$ from (\ref{uniformphasedistrib}),
and the kernel for LCC $K_{\ast_{\!B}}(d)$ equals the kernel for FPE with Hadamard product (\ref{fhp_kernel}), the sinc function.

\end{itemize}

\noindent
$\Box$

\subsection{Shaping the kernel in FPEs}
\label{sec:shapingFPEkernels}
  
\subsubsection{Phase distribution determines kernel shape}
By drawing the base vector of an FPE from distributions other than the uniform band-limited distribution (\ref{uniformphasedistrib}), one can design kernels with shapes that differ from the sinc function. The Bochner theorem (\ref{bochner}) states that any kernel whose Fourier transform is a proper density function can be represented with Fourier features drawn from this density \citep{rahimi2007random}. Equation (\ref{kerninner}) suggests how the Fourier density of some kernels can be used to construct an FPE. For each kernel that has a Fourier density $p(\omega)$ with support not exceeding the frequency range $[-1/2, 1/2]$, the phase density $p(\phi/(2 \pi) )/(2 \pi)$ is within the support $[-\pi, \pi]$, and generates random base vectors of an FPE with this kernel. 

\begin{figure}[H]
\centering
\includegraphics[width=1.0\columnwidth]{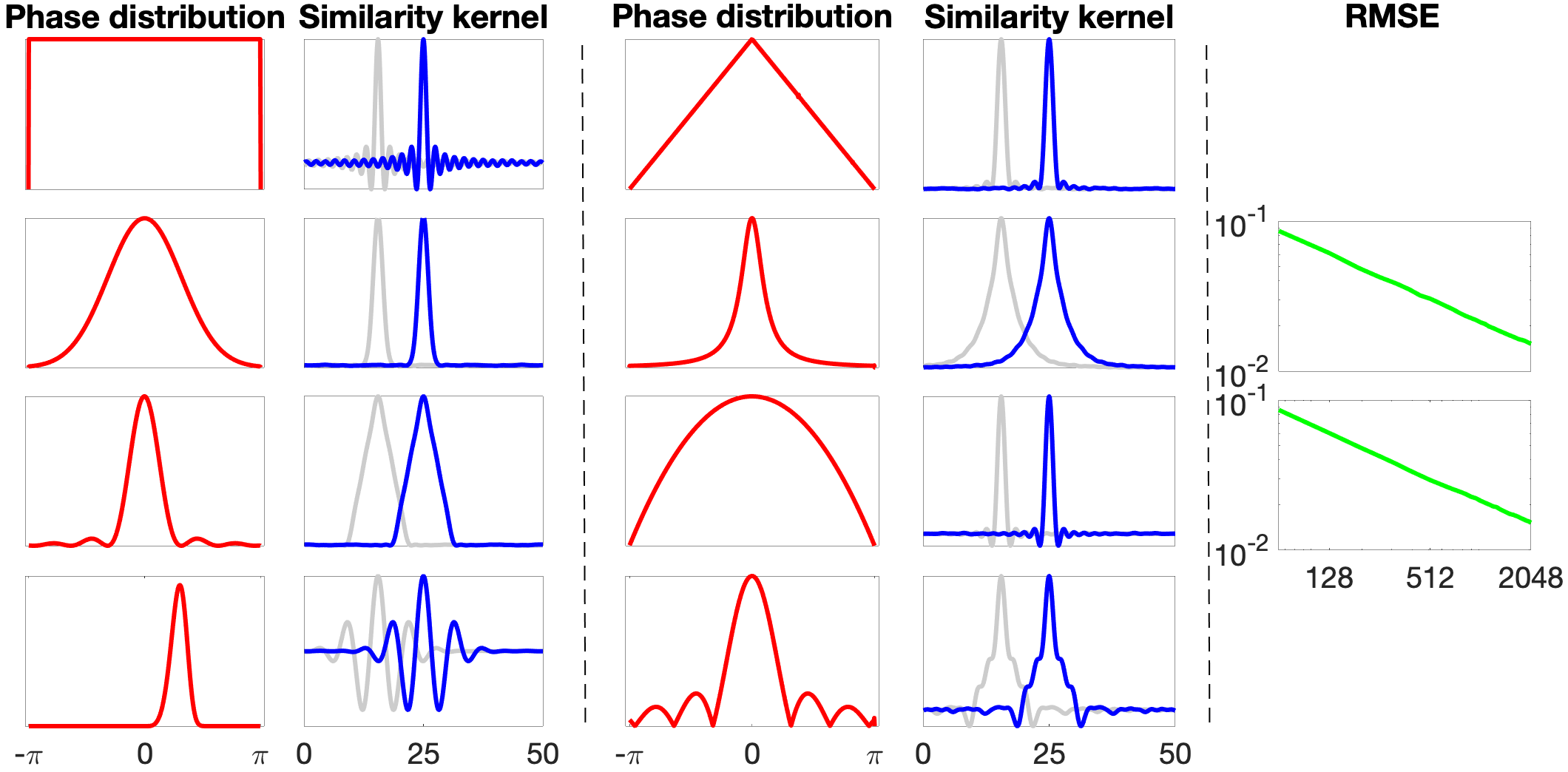}
\caption{
Similarity kernels for FPEs with base vectors sampled from various phase distributions and examples of RMSE convergence . 
Panels with red curves illustrate various probability density functions used to randomly sample phases for the base vector.
The corresponding panels with gray and blue curves show the similarity kernels resulting from the particular choice of phase distribution for $r=15.5$ and $r=25.0$, respectively. The similarity kernels are averages computed from $100$ simulation runs, $n=1,024$.
Panels with green curves depict the relation between the RMSE (ideal kernel vs. an FPE realization) and dimensionality of FPE vectors $n$ in the logarithmic scale.
The reported RMSE are averages computed from $100$ simulation runs; $n$ was in range $[64, 2048]$ with step $64$.
}
\label{fig:kernels}
\end{figure}

Figure~\ref{fig:kernels} shows several specific phase distributions (depicted in red) of base vectors, resulting shapes of the corresponding FPE kernels (in blue) for $n=1,024$, and convergence of Root Mean Square Error (RMSE) between the ideal kernel and its concrete FPE realization for increasing values of $n$ (logarithmic scale). Figure~\ref{fig:kernels} shows the RMSE for only a couple of kernels since all of them demonstrate the same dependency. 

The upper left panel depicts the FPE with uniform distribution and the sinc function kernel. The other panels show that it is possible to get similarity kernels, which are very close to, e.g., Gaussian (left panels in the second row), Laplace (right panels in the second row), or triangular (left panels in the third row) kernels. 
In fact, the shape of the kernel can be very complex such as the one in the lower right panels where the distribution of phases was defined using the truncated absolute values of the sinc function. 
Moreover, the RMSE between the ideal kernel and its FPE approximation decays exponentially with $n$.

\subsubsection{Periodic kernels}
\label{sec:perkern}
\begin{figure}[H]
    \centering
    \includegraphics[width=0.6\textwidth]{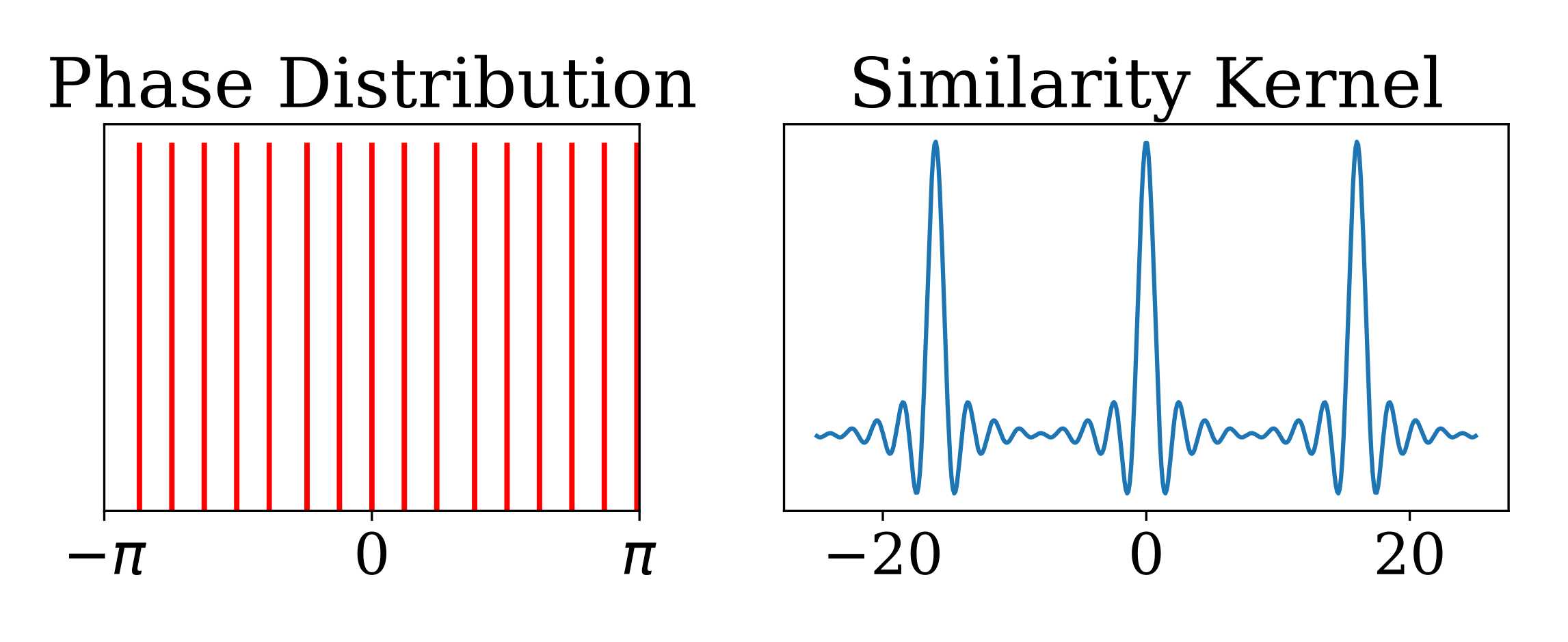}
    \caption{Periodic kernels are formed by sampling phases from the discrete roots of 1. The periodicity of the similarity is based on the order of the root.}
    \label{fig:periodic_kernel}
\end{figure}

It is clear from the Bochner theorem that periodic kernels will correspond to a discrete Fourier transform. 
Generally, the periodicity of the kernel is determined by the discrete sampling of the roots of the identity vector for each binding operation. 
In the case of Hadamard binding, periodic kernels result from phasor base vectors in which each component is sampled independently from a discretized phase distribution. 
For instance, a kernel with periodicity of $l$ results from a base vector whose individual phases are uniformly sampled from the $l$ phase values, $z_i \sim \mathcal{U}\{e^{i 2\pi j / l}, \  \forall j \in \{1,...,l\} \}$. Data on a circular manifold can be encoded by a base vector with phase values that equal $l$-th roots of $1$ (see Figure~\ref{fig:periodic_kernel} for an example).
These properties extend to circular convolution binding by sampling the discrete phases in the Fourier domain.
The LCC code produces loops when the hot phasor elements are also discretely sampled. The looping behavior depends more complexly on the products of common factors depending on block size and phase discretization. 
For instance, if the block code is purely binary (all phases are 0), then the cycle period will be the block size. 

\subsection{Multi-dimensional FPE formed with binding}
\label{sec:multidimensional:FPEs}

FPE can be extended to encode multi-dimensional data spaces:
\begin{equation}
    f_{\mathrm{LPE}}: \mathbf{r}\in \mathbb{R}^m \to \mathbf{z}(\mathbf{r}) \in \mathbb{C}^n \mbox{ where } m>1 
\end{equation}
A simple method for a multi-dimensional FPE is a Cartesian combination of one-dimensional FPEs that has been proposed before \citep{WeissOlshausenSpatial16,fradyframework18}.  
The Cartesian construction involves orthogonalization of the individual FPEs with the binding operation: 
\begin{equation}
    \mathbf{z}(\mathbf{r}) = \mathbf{z}_{1}(r_1) \circ \mathbf{z}_{2}(r_2) \circ \cdots \circ \mathbf{z}_{m}(r_m)
    \label{highdim_encode}
\end{equation}
with $\mathbf{z}_{i}(r_i)$ being one-dimensional FPEs independently sampled. It is important that the different FPEs are mutually independent, that is, the (unlikely) case has to be excluded that there are any intersecting paths.

Using the encoding scheme (\ref{highdim_encode}), the inner product kernel $\mathbf{z}(\mathbf{x})^{\top}\overline{\mathbf{z}(\mathbf{y})} \to K(\mathbf{x}-\mathbf{y})$ is a valid inner product kernel (see Theorem 3 for Hadamard FPE in Supplement~\ref{sec:mdim_hadamard-fpe}). This can be generalized to the other binding operations considered in this paper via Parseval's theorem.
The functions represented in the resulting VFA are of the form $f(\mathbf{s}) = \sum_k \alpha_k K(\mathbf{r}^k-\mathbf{s})$ are represented by vectors 
$f \equiv \mathbf{y}_f$: 
\begin{equation}
    \mathbf{y}_f = \sum_k \alpha_k 
    \mathbf{z}_{1}(r^k_1)\circ \mathbf{z}_{2}(r^k_2) \circ \cdots \circ \mathbf{z}_{m}(r^k_m)
    \label{mult_function_rep}
\end{equation}
The operations in function space are the same as for the one-dimensional VFA. 

\begin{figure}[H]
\centering
\includegraphics[width=\columnwidth]{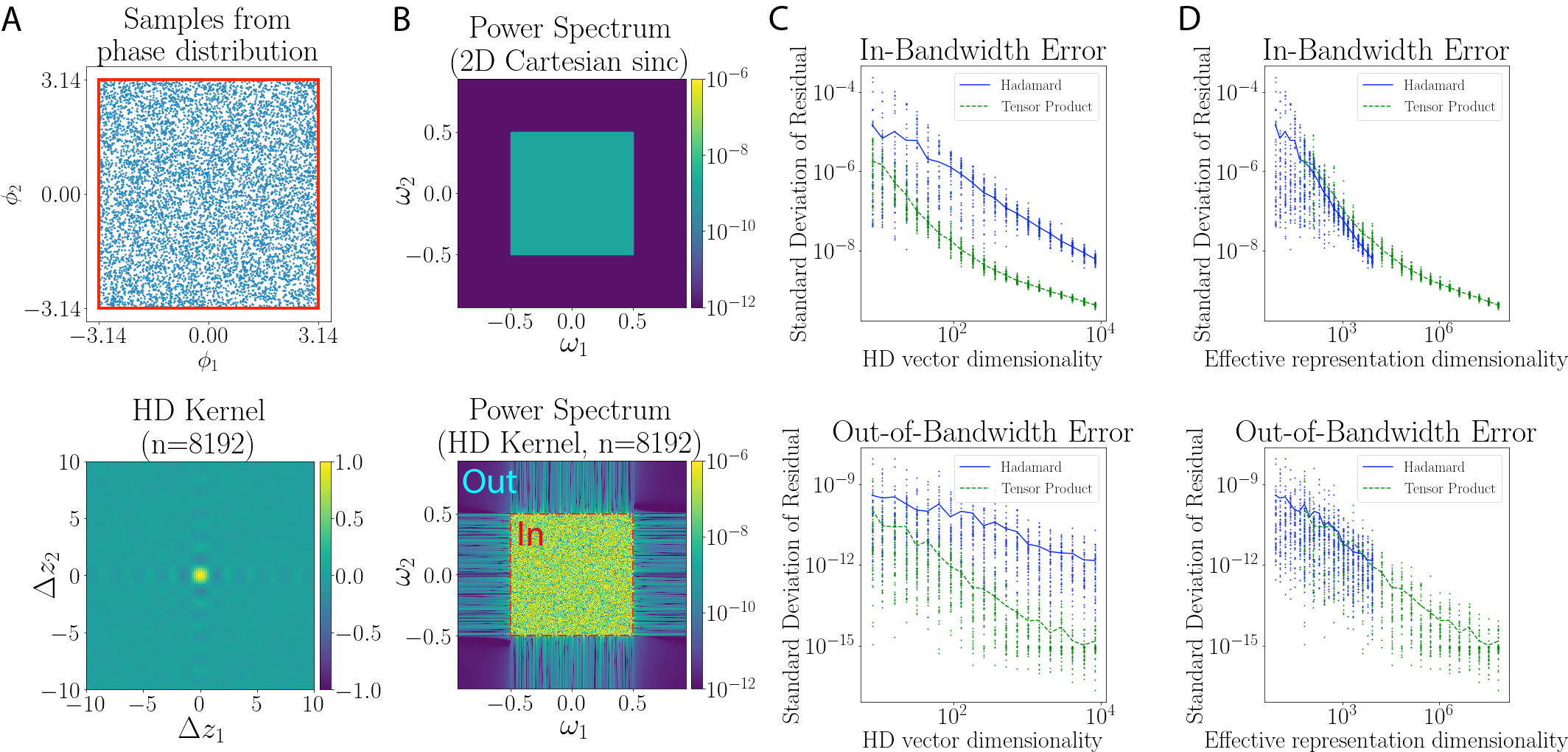}
\caption{ Convergence of binding of two FPEs to the 2-D Cartesian sinc kernel. A) Top: Scatterplot of phases drawn from $[-\pi,\pi]$. Each blue point is a pair of phases, and the red box denotes the interval boundary. Bottom: Resulting kernel. B)
Top: power spectrum of the 2-D Cartesian sinc kernel. Bottom: power spectrum of a kernel formed by Hadamard binding of two FPEs (\ref{highdim_encode}). With high enough dimension, the kernel approximates the 2-D Cartesian sinc kernel. Power spectra are normalized by number of samples. C) Higher dimensional representations show less in-bandwidth (top) and out-of-bandwidth error (down), respectively. Since the Fourier transform of a multi-dimensional sinc kernel is rectangular, in-bandwidth frequencies are defined as those frequencies inside the rectangle. Each plot shows the standard deviation of the difference between the power spectra of the 2-D sinc kernel and the FPE encoding. The blue trace is calculated for combination of the two dimensions by Hadamard product, the green trace for combination by the tensor product (\ref{smolensky_hadamard_fpe}) in Supplement~\ref{sec:mdim_hadamard-fpe}. D) Same as C, except plotting the dimension of the vectors after binding on the x-axis instead. The Hadamard product is a dimension-preserving binding, whereas the tensor product of two vectors results in an $n^2$-dimensional representation. In this regard, the Hadamard and tensor products have roughly equal convergence.
}
\label{fig:2d_cartesian_sinc}
\end{figure}

In the limit of large VSA dimension, the inner product of an FPE formed from multiple one-dimensional FPEs with sinc function kernels converges to the multi-dimensional Cartesian sinc function: 
\begin{equation}
    K_{sinc}(\mathbf{x}- \mathbf{y}) := K_{sinc}(x_1-y_1)\times \cdots \times  K_{sinc}(x_m-y_m)
\end{equation}
whose 2-D Fourier transform is a square. In Figure~\ref{fig:2d_cartesian_sinc} the combination of FPEs by binding (\ref{highdim_encode}) is compared to a combination by Smolensky's tensor product (\ref{smolensky_hadamard_fpe}) in Supplement~\ref{sec:mdim_hadamard-fpe}. Because Hadamard binding can be viewed as a lossy compression of the tensor product \citep{frady2020variable}, it is not surprising that the Hadamard product method converges somewhat more slowly. However, when comparing the methods on representations of the same dimensionality, the Hadamard product method converges slightly faster (within sinc bandwidth), see Figure~\ref{fig:2d_cartesian_sinc}D. 
Note that for finite VFA dimension, the largest deviations in power occur within the bandwidth of the sinc kernel, thus, high-frequency distortions and aliasing effects remain relatively small.

For concrete formulae describing a Cartesian multi-dimensional FPE with respect to Hadamard binding, see Section~\ref{sec:mdim_hadamard-fpe} of the Supplement.

\subsection{Shaping the kernels of multi-dimensional FPEs}

Multi-dimensional kernels can be shaped by shaping the one-dimensional kernels that compose them, as described in Section~\ref{sec:shapingFPEkernels}. Further, one can produce non-Cartesian kernels by sampling the base vectors of the FPEs encoding different dimensions from a joint phase distribution which does not factorize.

\subsubsection{Kernels with non-Cartesian structure}
\begin{figure}[H]
\centering
\includegraphics[width=\columnwidth]{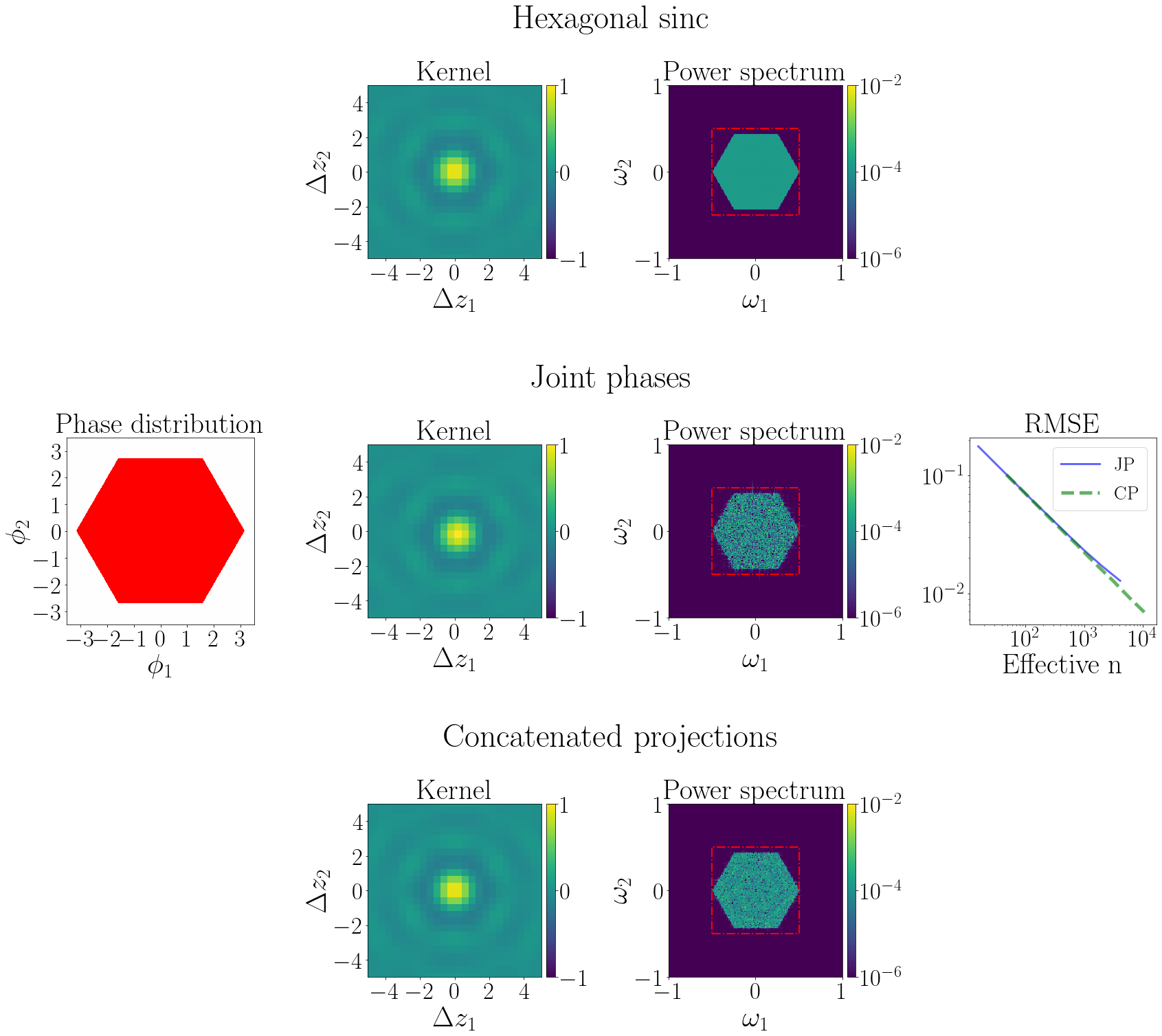}
\caption{
Top row: Hexagonal sinc function and its power spectrum. Second row: a method for generating hexagonal phases based on joint sampling of base vector phases uniformly from a hexagon inscribed in the valid phase range $[-\pi,\pi]$. This figure also converges to a hexagonal sinc function. Bottom row: Method for constructing phases based on 3 vectors with phases drawn from a uniform distribution. RMSE shows convergence to sinc function for both as a function of effective dimension ($n$).  For $n$-dimensional base vectors, the effective dimension is $n$ for the joint phases (JP) method and $3n$ for the concatenated projections (CP) method. Traces show average RMSE over 20 trials. Examples from the bottom two rows show one random example ($n=5,000$). Power spectra are normalized such that total power sums to 1.
}
\label{hexkern_pic}
\end{figure}

Like in the case of one-dimensional FPEs in Section~\ref{sec:shapingFPEkernels}, the Bochner theorem can be used to shape multi-dimensional kernels.
For example, one can produce FPEs with non-Cartesian sinc kernels. 
Sinc kernels can be generalized as the Fourier transform of the characteristic function of the Brillouin zone, a primitive cell of a Bravais lattice in reciprocal space \citep{ye2011geometric}. For example, the hexagonal sinc kernel (see Figure~\ref{hexkern_pic}), can be composed from three 2-D Cartesian sinc kernels as:
\begin{equation}
    K(\mathbf{x}) = \mbox{sinc}_{hex}(\mathbf{x}) = \frac{1}{3}\sum_{i=1}^3 \cos(\pi {\boldsymbol \xi}_i^{\top} \mathbf{x})\;\mbox{sinc}({\boldsymbol \xi}_{i+1}^{\top} \mathbf{x}) \;\mbox{sinc}( {\boldsymbol \xi}_{i+2}^{\top} \mathbf{x})
    \label{eq:hexkern}
\end{equation}
where the indices are modulo 3, and $\{{\boldsymbol \xi}_i:i=1,2,3\}$ are the vectors along the axes of the hexagonal grid: ${\boldsymbol \xi}_1=(1/4, \sqrt{3}/4)$, ${\boldsymbol \xi}_2=(1/4, -\sqrt{3}/4)$, ${\boldsymbol \xi}_3=(-1/2, 0)$. 
Since the Fourier transform of the hexagonal sinc function is the characteristic function of a hexagon, we can form the hexagonal sinc function by sampling pairs of phases of two base vectors to lie within a hexagon (Figure \ref{hexkern_pic}, middle row). With increasing dimension $n$, this method converges to the hexagonal sinc.


An alternate method for generating a hexagonal sinc function comes from using three base vectors with phases drawn uniformly from $[-\pi,\pi]$. The projections of the 2-D vector $\mathbf{x}$ on the three hexagonal grid vectors we call  $x_i = {\boldsymbol \xi}_i^{\top} \mathbf{x}$.
The encoding function:
\begin{equation}
    \mathbf{z}(\mathbf{x}) = \left(
    e^{\imath \pi x_1} \mathbf{z}_1(x_2) \circ \mathbf{z}_2(x_3), \ 
    e^{\imath \pi x_2} \mathbf{z}_1(x_3) \circ \mathbf{z}_2(x_1), \ 
    e^{\imath \pi x_3} \mathbf{z}_1(x_1) \circ \mathbf{z}_2(x_2)
    \right)^{\top}
\label{concatenatedprojections}
\end{equation}
produces an encoding vector that is a concatenation of three vectors formed from the representations of 2-D FPEs with the pseudo-orthogonal base phasor vectors $\mathbf{z}_{1}$ and $\mathbf{z}_{2}$. Using this encoding function, the real part of the inner product between two representations $\mathbf{z}(\mathbf{x})^{\top} \overline{\mathbf{z}(\mathbf{y})}$ resembles hexagonal similarity kernel (\ref{eq:hexkern}), $\mbox{sinc}_{hex}(\mathbf{x}-\mathbf{y})$. Figure~\ref{hexkern_pic} demonstrates that with increasingly higher dimensional vectors, equation (\ref{concatenatedprojections}) slowly converges to the hexagonal sinc function. This happens because the kernels of the 2-D FPEs are slowly converging to the 2-D Cartesian sinc kernel. For further discussion with the specific example of Hadamard binding, see equation (\ref{eq:frobenius_inner_product}) in the Supplement~\ref{sec:hex:other}.

\subsubsection{Non-Cartesian kernels with periodic structure}
\label{sec:multi_dim_period_kern}
\begin{figure}[H]
\centering
\includegraphics[width=\columnwidth]{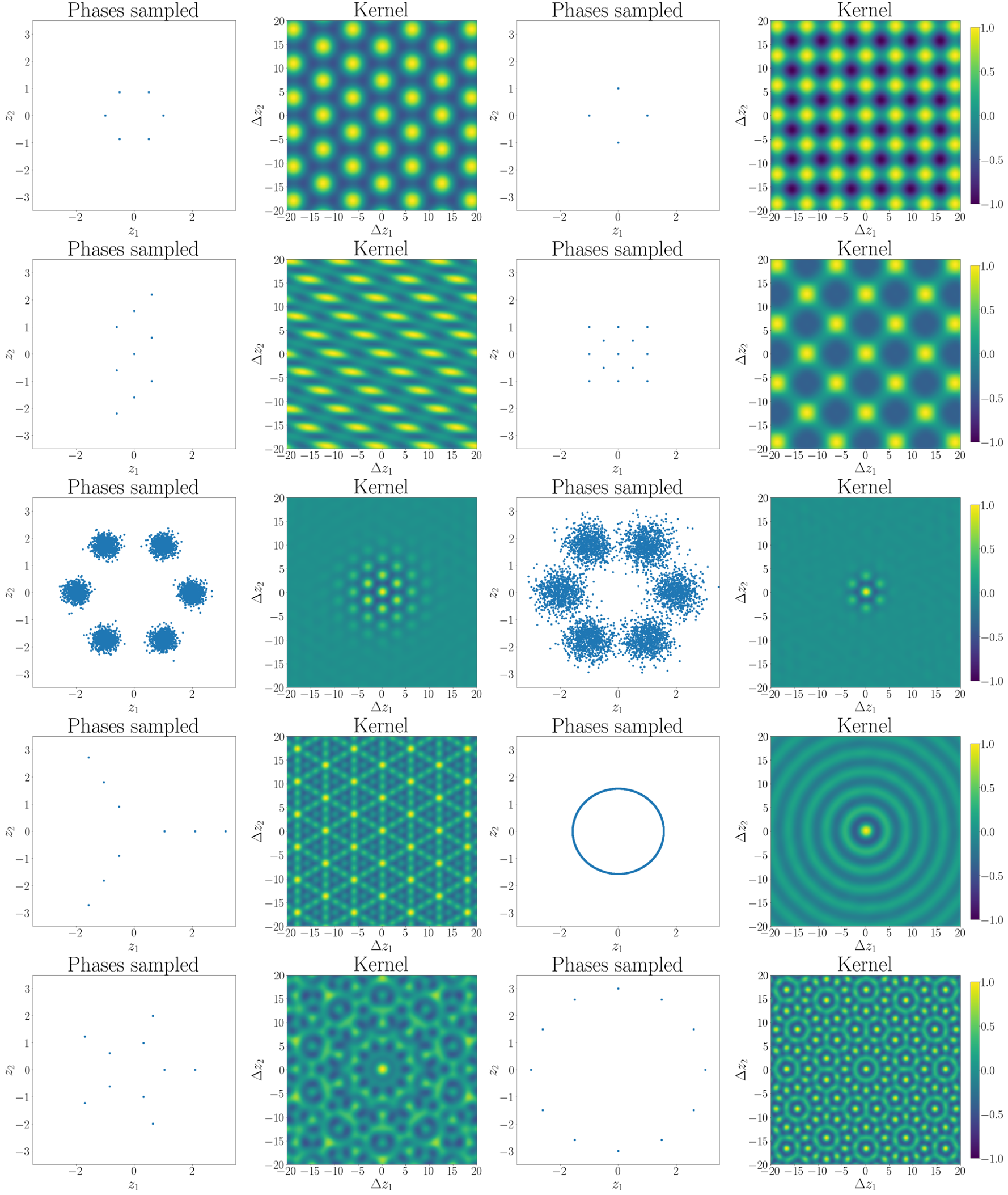}
\caption{
Examples of hexagonal and non-hexagonal lattices generated by strategic selection of phases jointly. Top two rows: Choosing phases of two base vectors jointly from the reciprocal of a 2-D Bravais lattice generates a kernel shaped in the form of that infinite 2-D lattice. The grid spacing of the kernel is determined by the spacing of the phases. Third row: When phases are chosen from an average of Gaussian distributions (whose centers form a reciprocal lattice) instead, the inner product goes to 0 with large distance. Left image is from Gaussians with $\sigma=0.2$, right is from Gaussians with $\sigma=0.4$. Higher variance results in a smaller window of non-zero inner product. Bottom two rows: Further examples of symmetric kernels that can be generated from a discrete phase distribution. All kernels are the result of 1 simulation with two random base vectors ($n=5,000$).
}
\label{lattices_kernels}
\end{figure}

Kernels with periodic structure play an important role in engineered and neurobiological navigation systems.
For instance in neuroscience, grid cells are a widely observed phenomenon in recordings from entorhinal cortex \citep{hafting2005microstructure}, where neurons fire at regular spatial intervals as the animal explores an environment, forming a hexagonal lattice.
To construct a hexagonal lattice in VFA, one can leverage concepts from crystallography. The molecules of a crystal can be described by a Bravais lattice, ideal grids constructed by a set of base vectors multiplied with integers. The Fourier transform of a Bravais lattice is called the reciprocal lattice, and the reciprocal lattice of typical Bravais lattices is band-limited and discrete. Thus, we can once again use the Bochner theorem and the relabeling of frequency with phase to construct LPEs with ideal grid-cell-like patterns. Indeed, the hexagonal lattice (Figure~\ref{lattices_kernels}, top left) is just one type of 2-D Bravais lattice. Other Bravais lattices can be constructed by choosing phases lying along their reciprocal lattice; for examples, see the first two rows in Figure~\ref{lattices_kernels}.  

Note that regular lattices are just a subset of discrete band-limited functions, the bottom two rows of Figure~\ref{lattices_kernels} illustrate other possibilities.

\begin{figure}[H]
\centering
\includegraphics[width=0.7\columnwidth]{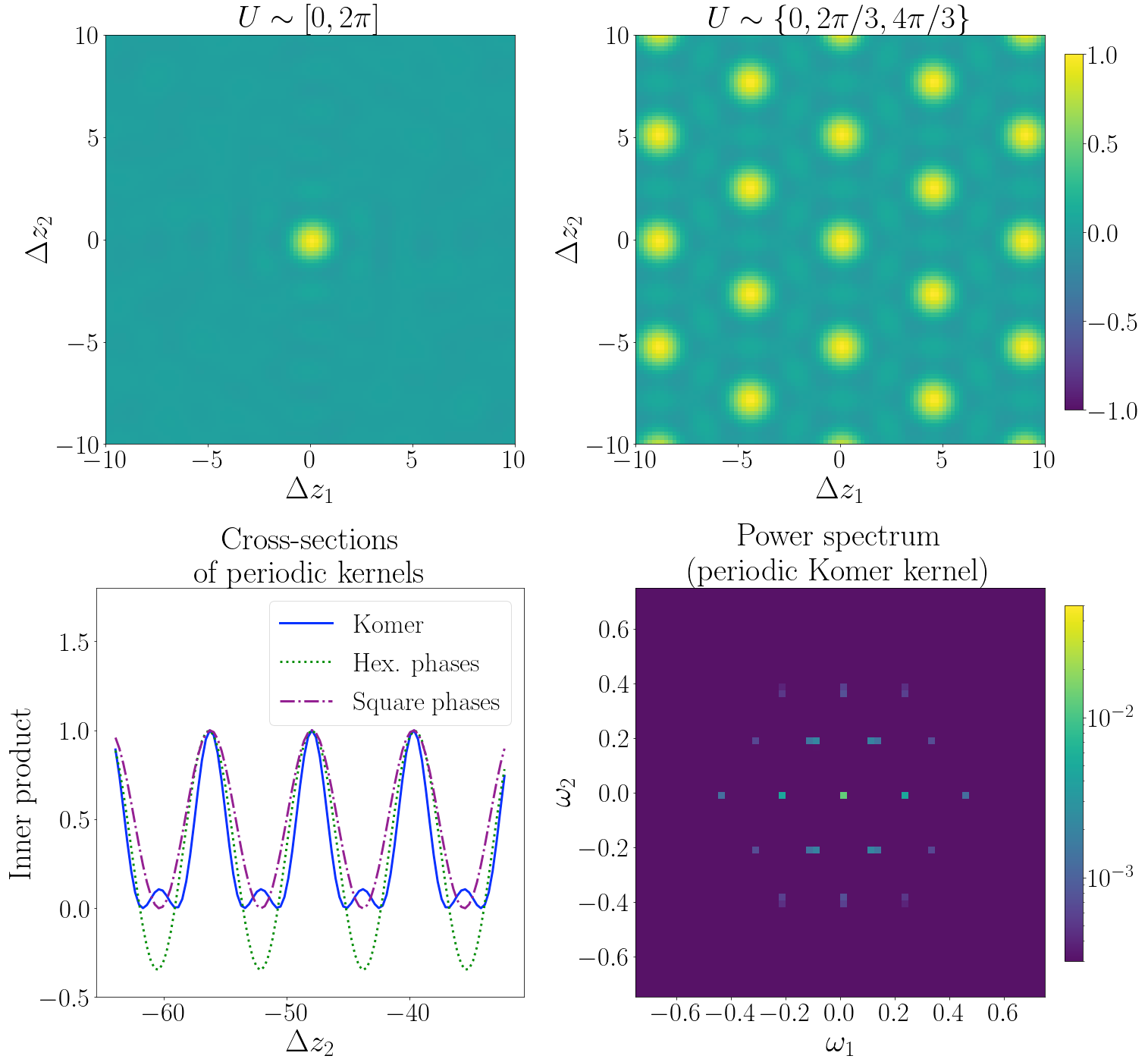}
\caption{Sampling discrete phases, as in \citep{komer2020biologically}, also results in a periodic kernel. Top left: Hexagonal kernel generated according to (\ref{twobasehexagon}), with phases for base vectors sampled uniformly from $[0,2\pi]$. This kernel is not periodic. Top right: Hexagonal kernel where phases are uniformly drawn from $\{0,2\pi / 3,4\pi / 3\}$ results in a periodic kernel ($n=5,000$). Bottom left: A cross-section of the periodic kernel shown in the top right. Cross-sections of kernel generated from (\ref{twobasehexagon}) have irregular structures between periods, unlike the hexagonal or square lattices in Figure~\ref{lattices_kernels} that  do not tend to have local maxima between peaks. Bottom right: The power spectrum of the periodic kernel (shown in top right) is band-limited and discrete yet also has higher power at lower frequencies.}
\label{hex_wrap}
\end{figure}

Our method is similar, but not identical, to the method for generating periodic kernels proposed in \citet{komer2020biologically}. Their method for generating grid-cell-like kernels is based on sampling discrete phases for the base vectors used in equation (\ref{komerformula_cc}). An example of this method is shown in Figure~\ref{hex_wrap}, and since its Fourier transform is also band-limited and discrete, its lattice appears quite similar to the hexagonal lattice we show in Figure~\ref{lattices_kernels}. Nevertheless, there are subtle differences. For example, using the method of \citet{komer2020biologically} generates local maxima in-between peaks of cross-sections of the kernel, whereas our method of choosing phases jointly does not create these additional local maxima.\footnote{Another difference is that the method of \citet{komer2020biologically} requires sampling from more discrete phases to change the grid scale, whereas our method based on the Bochner theorem only requires changing the spacing of the phase pairs).}

\section{Detection, decoding and denoising in VFA}
\label{sec:VFA:detection}
Function manipulation in VFA requires methods for detecting and decoding encoded values and functions. In this section, we describe a circuit model for detection and decoding in VFA. 
Recall from Section~\ref{symbolic_VSA_kernel} that in symbolic VSA, substantial computation is devoted to error correction, i.e., the detection, decoding, and denoising of VSA representations \citep{KleykoComputingParadigm2021}. 
A similar procedure for detection and error correction has to be part of VFAs. In a VFA, only the vectors in the subsets $\mathbf{P}_{\mathbf{z}}$ and $\mathbf{Y}_{\mathbf{z}}$ in Figure~\ref{fig:fvsa_visu} have meaning in the context of the function space. We explain first the group theoretic structure of $\mathbf{P}_{\mathbf{z}}$. Then we describe the proposed method, and a circuit for computing it, in the context of a Hadamard FPE encoding a one-dimensional data manifold. 

\begin{figure}[H]
\begin{center}
 \begin{tikzpicture}[fill=gray]
\scope
\clip  
(1,0) circle (1);
\fill[orange] (0,0) circle (1);
\endscope
\scope
\clip (-2,-2) rectangle (2,2)
      (0,0) circle (1);
\fill[green] (1,0) circle (1);
\endscope
\draw (1,0) circle (1) (1.5,0.5)  node [text=black,below] {$\mathbf{Y}_{\mathbf{z}}$}
      (0.5,0.28) ellipse (90pt and 45pt)
(0.5,1.6) node [text=black,below] {Entire VFA vector space};
\coordinate (A) at (0.4,-0.5);
\coordinate (B) at (0.6,-0.4);
\coordinate (C) at (0.4,-0.0);
\coordinate (D) at (0.8,0.4);
\coordinate (E) at (0.5,0.8);
\draw (-0.4,0.5) node [text=black,below] {$\mathbf{P}_{\mathbf{z}}$};
\draw plot [smooth] coordinates {(A) (B) (C) (D) (E)};
\draw (A) node [text=black] {$\bullet$} (B) node [text=black] {$\bullet$} (D) node [text=black] {$\bullet$} (D) node [text=black] {$\bullet$} (E) node [text=black] {$\bullet$}; 
\draw (A) node [text=black,above] {$\mathbf{z}$} (0.8,-0.4) node [text=black,above] {$\mathbf{z}^{\beta}$} (D) node [text=black,left] {$\mathbf{z}^{2\beta}$} (E) node [text=black,right] {$\mathbf{z}^{3\beta}$};
\end{tikzpicture}
\vspace{-0.9cm}
\caption{The two matching steps of the proposed decoding method visualized in the state space. Decoding in a VFA serves to determine whether and where an unkwown vector lies in the set of interpretable vectors, depicted by the circle. For a representation of an unknown point on the data manifold, the coarse matching step compares the vector to a set of discrete anchor points, marked by $\bullet$, on the path of point representations, symbolized by the line. Once the best coarse match is found, the second decoding step determines the exact location of the point on the path between the anchor points. If the vector encodes a function, i.e., lies in the green region, the same two matching steps are repeated multiple times, each time decoding the coefficient and support point of one term in the function (\ref{expand}). One term, once decoded, is subtracted from the input vector before the decoding process is repeated until all terms constituting the function are determined.}
\label{fig:decode_visu}
\end{center}
\end{figure}
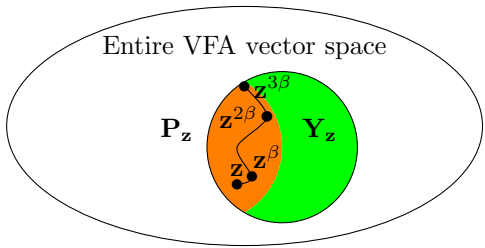

For the design of a decoding procedure it is important to understand the group theoretic structure of the set of point representations $\mathbf{P}_{\mathbf{z}}$. 
There is a direct relation between the Hadamard FPE and the exponential map, $t \to e^{H t}$, of diagonal skewed Hermitian matrices $H$, which produces a subset of unitary matrices. 
For a particular FPE with base vector $\mathbf{z} = (e^{\imath \phi_1},...,e^{\imath \phi_n})$, consider the exponential map of a diagonal skew Hermitian matrix $H(\mathbf{z}) = \imath \;\mbox{diag}({\boldsymbol \phi})$, with a real-valued diagonal matrix formed from the phase angles of $\mathbf{z}$. 
The exponential map takes a single $H(\mathbf{z}): \mathbf{z} \in A_{\odot}$ as the generator, and forms the subgroup by the so-called ``path'', a one-parameter manifold of matrices, a Lie group \citep{gilmore2012lie}. It is known that the exponential map of a general skew Hermitian matrix forms an injective map onto the unitary matrices \citep{gallier2020differential}. Thus, the path formed by the single point representations in a Hadamard FPE is a one-parameter Lie group which is isomorphic to a subgroup of the unitary matrices. This parametrized group structure of $\mathbf{P}_{\mathbf{z}}$ is symbolized by the line in Figure~\ref{fig:decode_visu}. Due to the non-uniqueness of the complex exponential, for certain base vectors the path can also be a loop, that is, periodically revisit the same unitary matrices. This loop structure yields periodic kernels 
as seen in Section~\ref{sec:perkern}. 

For an unknown given vector, the decoding problem consists of decomposing the vector into a linear combination of vectors on the path $\mathbf{P}_{\mathbf{z}}$.
If vectors are close  are only near the path they should be error corrected to lie on the path and defining the encoded value. If the vector cannot be decomposed into points near the path, the decoding should be rejected.

The nature of the decoding problem in VFA is not a new one. In neuroscience, line attractor networks have been proposed as models for how a neural circuit can decode, denoise, and represent real-valued inputs. In principle, we could use a line attractor in VFA. This would require a circuit that can stabilize and denoise the entire path for an FPE with a given base vector. 
However, it is unclear how to store the path of a specific FPE in a line attractor network. 
A naive solution could be to finely discretize the path and store the resulting points in a Hebbian associative memory. But this might fail because a sufficiently fine path discretization could result in a set of patterns, which is too large and its members too correlated to be successfully stored in a Hebbian associative memory circuit \citep{lewenstein1992optimal}.

Thus, here we propose instead a two-step procedure for detection and error correction, including a coarse and fine matching steps. For coarse matching, we store anchor points on the path in a content-addressable memory, see the points on the line in Figure~\ref{fig:decode_visu}. For forming the coarse anchor points of the path,  the base vector exponentiated at an interval somewhat smaller than $2$, for example, $\{\mathbf{z}^{k \beta}: k=1,2,k_{max}, \beta = 2 \times 0.8\}$. This discretization of the path is chosen such that for a representation of any point on the path, the closest anchor point has a higher inner product compared to the other anchor points. The coarse matching step can be accomplished in a phase associative memory, in which all the anchor points are stored using the standard outer product learning rule \citep{frady2019robust}.  

\begin{figure}[H]
\centering
\includegraphics[width=1.0\columnwidth]{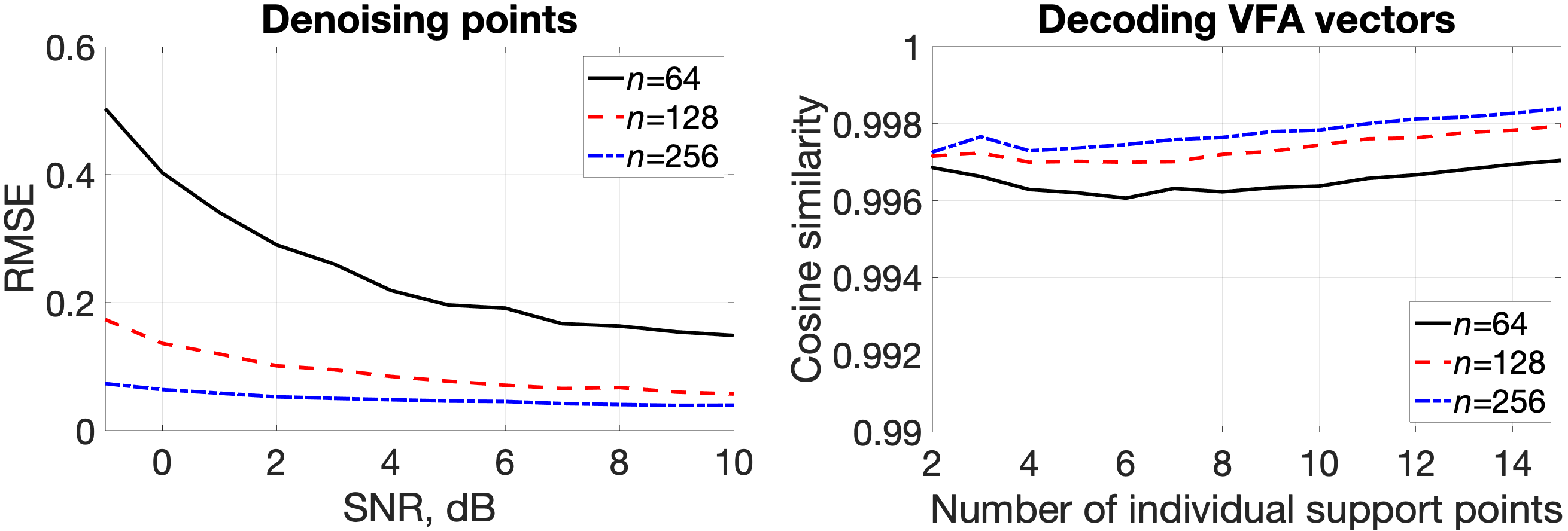}
\caption{
Decoding VFA representations with the circuit in (\ref{denoise_pass}).
For both panels $k_{max}=20$, $\beta = 2 \times 0.8$, $n \in \{ 64, 128, 256\}$.
Left panel: denoising point encoding vectors.
The results were averaged over $50,000$ simulations. 
Right panel: decoding of function vectors. 
The results were averaged over $5,000$ simulations. 
}
\label{fig:detection}
\end{figure}

Once the coarse matching step returns the closest anchor point, the second mechanism precisely decodes the analog value. 
This mechanism involves the FPE $\mathbf{z}(s)=\mathbf{z}^{s}$ also used to encode input vectors,  and a circuit forming the inner product between the representation of a value $s$ and the unknown vector $\mathbf{x}$.
\begin{equation}
    c(s) = [\mathbf{z}^s]^{\top}\overline{\mathbf{x}} = \sum_j e^{\imath(s-r)\phi_j}
    \label{denoise_pass}
\end{equation}
The phase distribution in the sum in (\ref{denoise_pass}) is centered rectangular (\ref{uniformphasedistrib}) with width $s-r$. 
Maximizing $c(s)$ yields the encoded value $s=r$, as this setting shrinks the width of the phase distribution to zero in all components simultaneously. 
Thus, if $\mathbf{x}$ is the representation of a single point $r$ ($\mathbf{x}=\mathbf{z}(r)=\mathbf{z}^r$), the control signal $c(s)$ changes like the kernel function as $s$ is approaching $r$. As long as the anchor points are close enough so that the kernel increases monotonously, the fine matching procedure is convex optimization and can be solved via, e.g., the gradient descent.
Left panel in Figure~\ref{fig:detection} depicts the RMSE resulting from the denoising over the input signal-to-noise ratio (SNR) for three different values of vector dimension $n$.

Decoding can also be accomplished for function vectors in the VFA, i.e., non-phasor vectors which are a weighted superposition of the FPE vectors of the support points of the function (\ref{eq:vfa:function}). To decode the weight value and support point with largest weight, one can just process the function vector in the same way as a single point encoding vector described above. Once the support point value $r_1$ is decoded, one can subtract the contribution of the first support point $\hat{\alpha}_1 \mathbf{z}^{\hat{r}_1}$ from the function vector and then proceed to decode weight value and support point with the second largest weight, and so on. 
Decoding of the function vectors is demonstrated in the right panel of Figure~\ref{fig:detection}.
The average cosine similarity between the original vector representation of a function and the representation of the decoded function is plotted over the number of individual support points in the VFA vector for three different values of $n$.


\section{VFA application examples}
\label{sec:kernel:methods}

We have shown so far that FPE induces a VFA in which vectors represent the band-limited functions. Here we describe concrete applications of VFAs of this type, for image processing (Section~\ref{sec:proc_im_dat}), density estimation (Section~\ref{sec:dens:est}), and  nonlinear regression (Section~\ref{sec:nonl:regr}). 

\subsection{Processing image data}
\label{sec:proc_im_dat}

We start by describing how to represent and manipulate image data by high-dimensional vectors, an example that showcases critical VFA properties. In essence, we treat simple images of letters as functions over the image domain. 
For simplicity, this demonstration uses the Hadamard VFA.
Locations in the image plane are encoded by a two-dimensional FPE, 
with base vectors $\mathbf{x}$ and $\mathbf{y}$.
To avoid a reduction of image power when translations move image parts outside the image boundary, 
we desire a function space with a torus structure that seamlessly connects the image boundaries. To achieve this, we shape a periodic kernel by choosing the base vectors from a discretized phase distribution as described in Section~\ref{sec:multi_dim_period_kern}. 
For images of size 56x56 pixels, the phases of $\mathbf{x}$ are chosen from the 56 discrete points around the phase circle, likewise for $\mathbf{y}$.

\begin{figure}[H]
    \centering
    \includegraphics[width=0.9\textwidth]{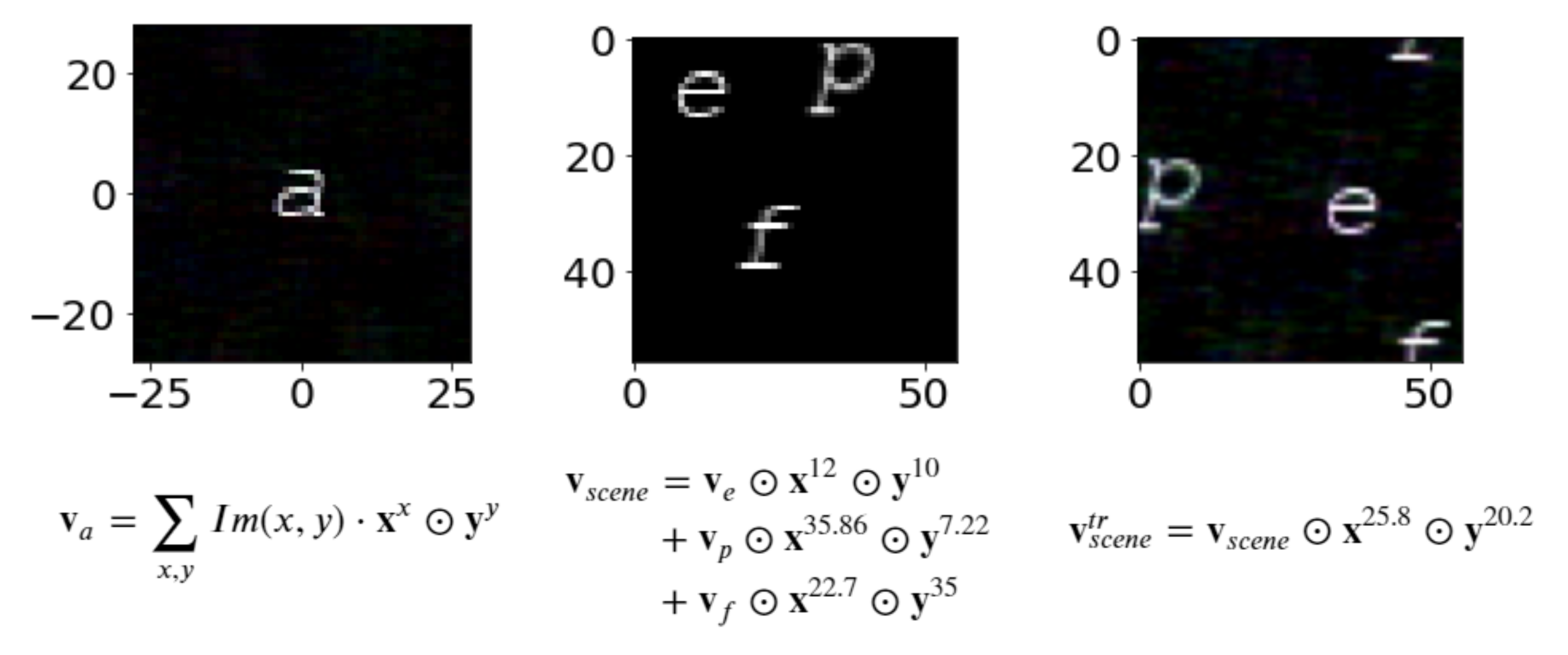}
    \caption{Encoding and manipulating images. Each panel shows an image decoded from a vector. Left: Images of letters are encoded as functions over the image domain. Each pixel is encoded by a support vector representing its location, weighted by the pixel intensity. A complete letter image is represented as the VFA vector containing the sum of all individual pixel terms. Middle: A scene of several letters can be generated by positioning and adding individual letter vectors. Right: The entire scene can also be manipulated by binding the scene vector with an FPE translation vector. To keep the image power invariant under translations, we use a VFA with periodic kernel. The kernel creates a torus topology, that is, when shifting the scene, parts that move outside the image frame wrap around and enter the image on the opposite side.}
    \label{fig:image}
\end{figure}

A letter image is an element of the VFA function space, with each term in the kernel expansion of the function (\ref{expand}) encoding intensity and position of a pixel by coefficient and support point, respectively, see left panel in Figure~\ref{fig:image}. 
Several letters can be composed into a scene by adding the individual image vectors (center panel in Figure~\ref{fig:image}). The letters can be positioned in the scene by binding each image vector with fractional power encoding of the position. 
The scene vector can then be decoded through a coarse search as described in Section~\ref{sec:VFA:detection}: 
\begin{equation}
    \hat{Im}(x, y) = \frac{1}{n}(\mathbf{v}_{scene})^\top \overline{(\mathbf{x}^{x} \odot \mathbf{y}^y)}
\end{equation}
The entire scene can be shifted by again binding the scene vector with FPE translation vectors. Due to the torus structure of the function space the letters wrap around the image. 

\subsection{VFA models for nonparametric kernel methods}
\label{sec:vfa_nonpara_ker_met}
\subsubsection{Density estimation}
\label{sec:dens:est}

The application of VFA in density estimation can leverage early work using sinc kernels in density estimation \citep{davis1975mean, davis1977mean, devroye1992note,glad2003correction,agarwal2015nonparametric}.

\paragraph{\textit{\textbf{Method}}}\mbox{}\\ 
For a given set of $k$ data points $r_1,...,r_k$, \citet{agarwal2015nonparametric} proposed a method for density estimation with band-limited functions. We use this method to construct a vector $\mathbf{y}^p$ that represents the band-limited probability maximum likelihood (BLML) estimator for the density $p(r)$ of the data points:
\begin{equation}
    p(r) = \left( \frac{f_c}{k} \sum_{i=1}^k  
    \hat{c}_i K(f_c(r-r_i)) \right)^2 
    = \left( \frac{f_c}{kn} \sum_{i=1}^k 
    \hat{c}_i \mathbf{z}(f_cr_i) \overline{\mathbf{z}(f_cr)} \right)^2 
    = \left( (\mathbf{y}^p)^{\top} \overline{\mathbf{z}(f_cr)} \right)^2
    \label{eq:dens:est}
\end{equation}
with $\mathbf{y}^p := \frac{f_c}{kn} \sum_{j=1}^k \hat{c}_j \mathbf{z}(f_cr_j)$, $f_c$ the cut-off frequency, and 
$\hat{\mathbf{c}} = \underset{\rho_n(\mathbf{c})=0}{\mbox{argmax}} \prod_{i=1}^k \frac{1}{c_i^2}$, $\rho_n(\mathbf{c})_i = \frac{1}{k} \sum_{j=1}^k c_j K(f_c(r_i-r_j)) - \frac{1}{c_i}$.
An equivalent formulation of the optimization to find $\hat{\mathbf{c}}$ is:
\begin{equation}
    \hat{\mathbf{c}} = \underset{\mathbf{1} = \mathbf{c} \odot \mathbf{K} \mathbf{c}}{\mbox{argmax}} \;\;\prod_{i=1}^k \frac{1}{c_i^2}
    \label{eq:BLML:c}
\end{equation}
where $\mathbf{1}$ is the $k$-dimensional all one-vector, and $\mathbf{K}$ is the Gram matrix of the representation vectors of the $k$ data points. This optimization problem involves exhaustive search among the $2^k$ local maxima (one in each orthant). To avoid this expensive search, \citet{agarwal2015nonparametric} proposed three approximate solution algorithms. In the experiments below, we used \texttt{BLMLTrivial} algorithm. 
If $p(x)$ is indeed band-limited and strictly positive then this algorithm converges to the true BLML estimator asymptotically.

\paragraph{\textit{\textbf{Empirical evaluation}}}\mbox{}\\ 
We use an example of a surrogate probability density function (pdf) from~\citet{agarwal2015nonparametric}:
\begin{equation}
    f(x)=0.078(\mbox{sinc}^2(0.2x)+ \mbox{sinc}^2(0.2x + 0.2))^2.
    \label{eq:surr:pdf}
\end{equation}
\noindent
This pdf is band-limited with a cut-off frequency of $f_c=0.4$ Hz.

We followed the experimental setup in~\citet{agarwal2015nonparametric} by first randomly drawing $k$ samples from the pdf and then using these samples to estimate the density of the pdf in the range $[-5,5]$ with step $0.001$. 
The values of $\hat{\mathbf{c}}$ were obtained with the \texttt{BLMLTrivial} algorithm~\citet{agarwal2015nonparametric}.
It is a computationally efficient  algorithm as it obtains $\hat{\mathbf{c}}$ in one step by first selecting an orthant in which the global maximum for (\ref{eq:BLML:c}) may lie. 
Once the orthant is chosen, $\rho_n(\mathbf{c})=0$, which is monotonic in a given orthant,  is solved in that orthant (see Section~3.3.2 in~\citet{agarwal2015nonparametric} for details).
We used the mean integrated square error (MISE) between estimate and ground truth value and an indicator of the quality of the estimation. 
\begin{figure}[H]
\begin{minipage}[h]{0.28\linewidth}
\center{\includegraphics[width=1.0\linewidth]{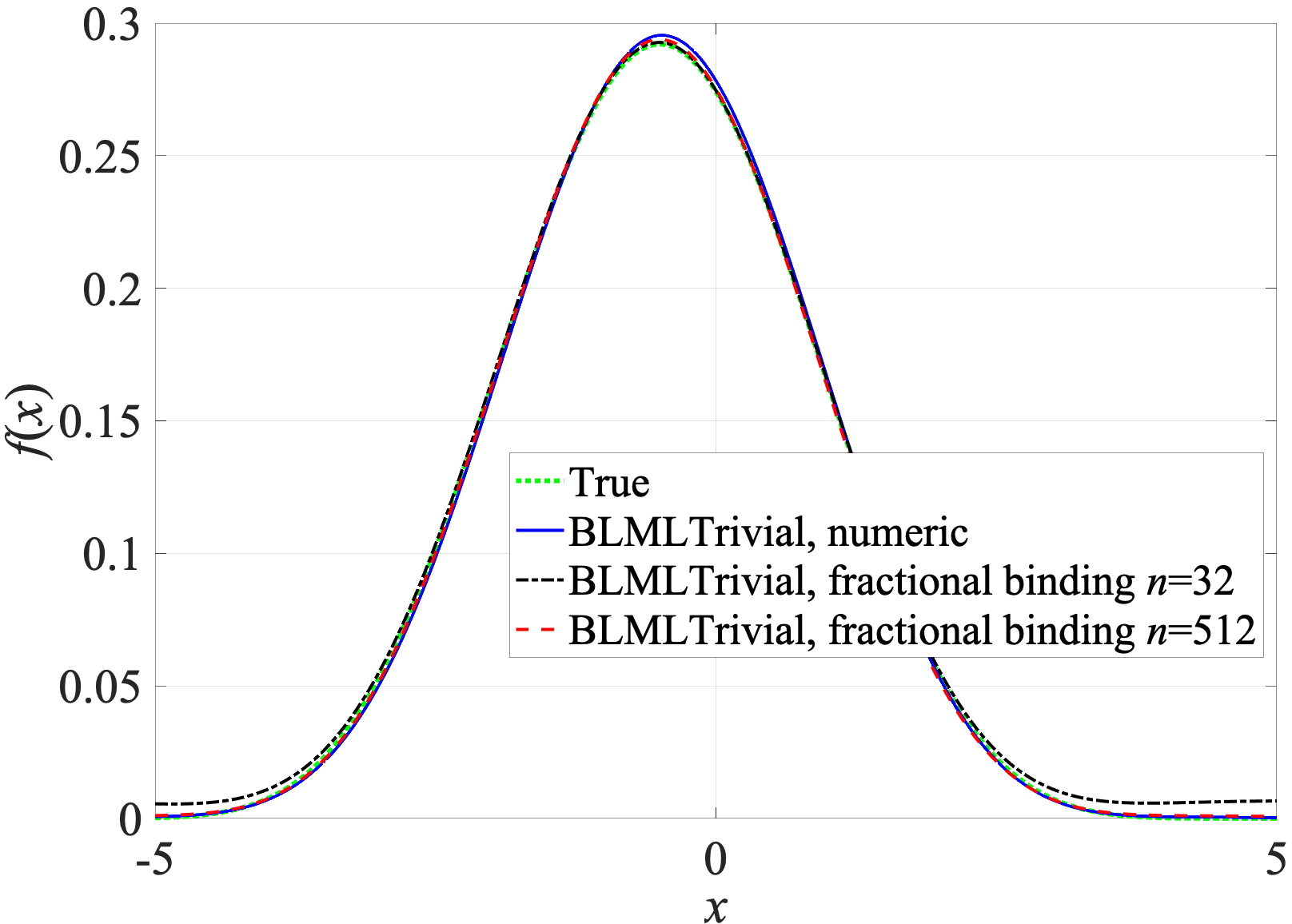} }
\end{minipage}
\hfill
\begin{minipage}[h]{0.28\linewidth}
\center{\includegraphics[width=1.0\linewidth]{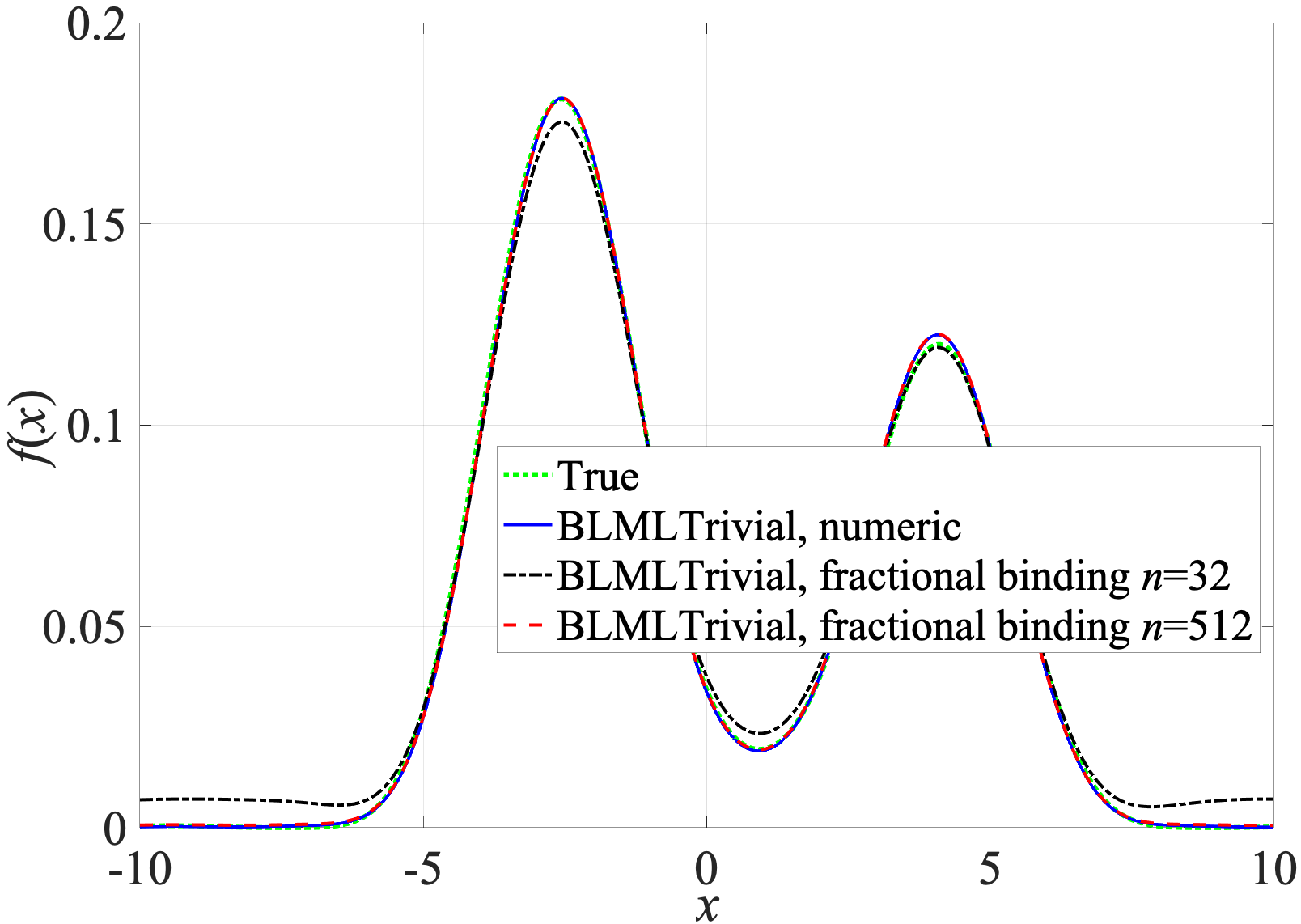} }
\end{minipage}
\begin{minipage}[h]{0.42\linewidth}
\center{\includegraphics[width=1.0\linewidth]{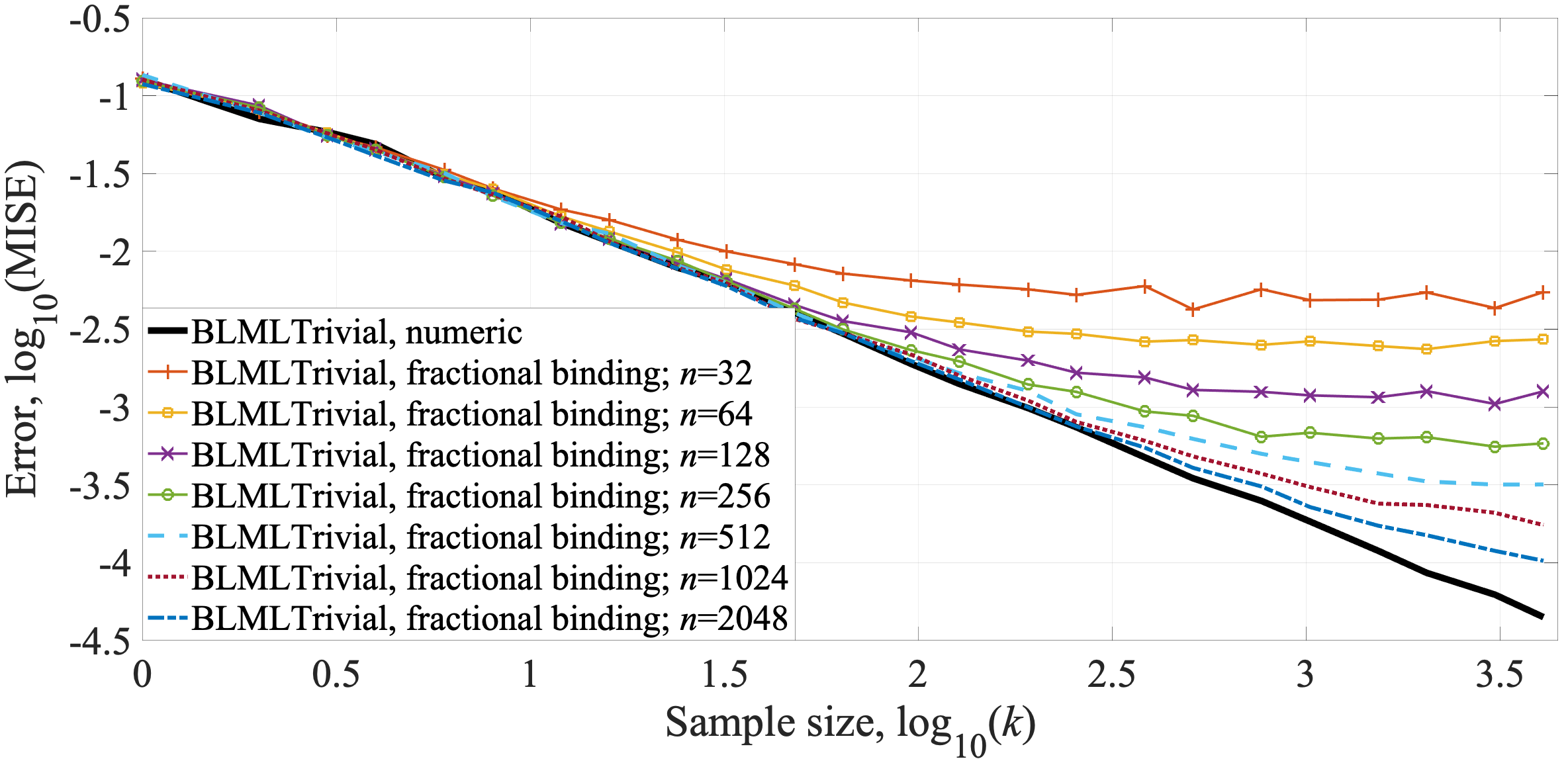} }
\end{minipage}
\caption{
Left panel: True pdf ((\ref{eq:surr:pdf}), dotted line) and estimations obtained with VFA implementing the \texttt{BLMLTrivial} algorithm. The number of samples from $f(x)$ was fixed to $k=81$. 
Center panel: True pdf -- $0.170(0.8\mbox{sinc}^2(-0.2x+0.8)+ \mbox{sinc}^2(0.2x + 0.5))^2$ (dotted line) and estimations obtained with VFA implementing different variants of the \texttt{BLMLTrivial} algorithm. 
In both panels, the solid lines correspond to the original variant from~\citet{agarwal2015nonparametric}; dashed and dash-dotted lines correspond to the proposed variant (\ref{eq:dens:est}) for different dimensionalities of the base vector $\mathbf{z}$. 
To be consistent with~\citet{agarwal2015nonparametric} (cf. Figure 1.B there) the number of samples was fixed to $k=81$. 
The reported pdf estimates are the mean values from $500$ simulation runs.
For each run, samples were drawn randomly from the corresponding pdf; 
for the variants with FPE, the base vector $\mathbf{z}$ was also chosen randomly at each run. 
Right panel: Logarithm of MISE plotted against the logarithm of the number of samples for different estimates of the pdf.  Numeric version of the \texttt{BLMLTrivial} algorithm (solid line) and estimates with VFA for seven different base vector dimensions. 
The number of samples varied in the range $2^{[0,11]}$ with step $0.5$.
MISE were averaged over $500$ simulation runs. For each run, samples were drawn randomly from (\ref{eq:surr:pdf}); for the VFAs, 
the base vector $\mathbf{z}$ was also chosen randomly at each run. 
}
\label{fig:pdf:example}
\end{figure}

Figure~\ref{fig:pdf:example} shows how the estimation quality of the VFA improves as the vector dimension is increased.
While a vector dimension of $n=32$ provides a decent estimate of the overall curve shape but overestimates the tails, for a vector dimension of $n=512$ the estimate is close to the ground truth even in the tails.
Thus, the dimensionality of the VFA affects the quality of the estimation. This observation enabled the next experiment, where both the number of samples $k$ and the dimensionality $n$ are varied jointly.


Note that in the right panel in Figure~\ref{fig:pdf:example}, the MISE of the numerical implementation of the \texttt{BLMLTrivial} algorithm (thick solid line) is steadily decreasing with the growing number of samples. This decrease reflects the improvement of precision as data limitation is gradually lifted. The MISE curves of VFA models follow this curve, but at some point deviate from the solid line and plateau. In practice this has the following consequences. 

In the regime of unlimited data, i.e., for large numbers of samples, the precision of VFA models steadily improves with its dimensionality. This improvement with dimensionality has also been observed in classical VSAs, and has been referred to as the ``blessing of dimensionality''~\citep{Gorban2018Blessing}. 

In the data-limited regime, i.e., for a small, fixed number of observation points, the situation is quite different. In this case, VFAs with large enough fixed dimension already achieve the same performance as the numerical variant. Increasing the dimensionality beyond this point does not improve the performance further.  Thus, in the data-limited regime, one can limit the dimensionality of the VFA to some fixed value without compromising performance. 

\subsubsection{Nonlinear regression}
\label{sec:nonl:regr}

To demonstrate nonparametric nonlinear regression with VFA, we can leverage some previous literature on regression with sinc kernels \citep{bissantz2007estimation, exterkate2011modelling, bousselmi2020reproducing}.

\paragraph{\textit{\textbf{Method}}}\mbox{}\\ 
The regression task is formulated as follows: given a training set $\{(X_i,Y_i), 1 \leq i \leq k \}$ with $k$ samples, where $Y_i$ is a noisy value of an unknown function $f(x)$ at $X_i$ ($Y_i = f(X_i) + \eta_i$); provide an estimate of the function for previously unseen values of $x$ -- $\hat{f}(x)$.

Two methods for performing regression with sinc kernels were recently described in~\citep{bousselmi2020reproducing}. 
The first method is called empirical projection, and it can be used in the special case when values of $X_i$ are independent and uniformly distributed on the interval $I=[-1,1]$ and $f(x)$ lies in a subspace of the Hilbert space $L^2(I)$. 
The empirical projection regression estimator with the sinc kernel is defined as: 
\noindent
\begin{equation}
    \hat{f}_{c,k}(x)= \frac{2}{k}\sum_{i=1}^k Y_i K_c(x, X_i) =\frac{2c}{k \pi} \sum_{i=1}^k Y_i \mbox{sinc}\left(\frac{c}{\pi}(X_i-x) \right),
\end{equation}
\noindent
where $c$ is the bandwidth parameter of the kernel. 

Using the FPE, the empirical projection method can be realized as follows: 
\noindent
\begin{equation}
    \hat{f}_{c,k,n}(x)= \frac{2c}{k \pi} \sum_{i=1}^k Y_i \mbox{sinc}\left(\frac{c}{\pi}(X_i-x) \right) = \frac{2c}{k n \pi} \sum_{i=1}^k Y_i \mathbf{z}(\frac{c}{\pi}X_i)^{\top} \overline{\mathbf{z}(\frac{c}{\pi}x)} = (\mathbf{y}^X)^{\top} \overline{\mathbf{z}(\frac{c}{\pi}x)},
\end{equation}
\noindent
where  $\mathbf{y}^X = \frac{2c}{k n \pi} \sum_{i=1}^k Y_i \mathbf{z}(\frac{c}{\pi}X_i)$ is the FPE-based estimator of $f(x)$.

The second method is Tikhonov regularization, which is more general as it does not require $X_i$ to be within $I$. The Tikhonov regularization regression estimator with the sinc kernel is defined as: 
\noindent
\begin{equation}
    \hat{f}_{c,k}^{\lambda}(x)= \sum_{i=1}^k c_{i, \lambda} K_c(x, X_i) =\frac{c}{\pi} \sum_{i=1}^k c_{i, \lambda} \mbox{sinc}\left(\frac{c}{\pi}(X_i-x) \right),
\end{equation}
\noindent
where $\lambda$ is a regularization parameter and $\mathbf{C}_{\lambda}= (c_{i, \lambda} )_{1 \leq i \leq k}$ is the expansion coefficients vector, which is calculated using regularized linear regression as: 
\begin{equation}
    \mathbf{C}_{\lambda}= G_{\lambda}^{-1} \mathbf{Y}= \left[ \left[ K_c(X_i, X_j) \right]_{1 \leq i,j \leq k} + k \lambda I_k \right]^{-1} \mathbf{Y}, \mathbf{Y}=(Y_i)_{1 \leq i \leq k},
    \label{eq:Tikhonov:expansion}
\end{equation}
where $I_k$ is the $k \cross k$ identity matrix and $\left[ K_c(X_i, X_j) \right]_{1 \leq i,j \leq k}$ is the Gram matrix for the training data $(X_i)_{1 \leq i \leq k}$.

Similar to empirical projection, the estimator for the Tikhonov regularization method can be expressed in VFA as: 
\begin{equation}
    \hat{f}_{c,k,n}^{\lambda}(x)= \frac{c}{\pi} \sum_{i=1}^k c_{i, \lambda} \mbox{sinc}\left(\frac{c}{\pi}(X_i-x) \right) = \frac{c}{n \pi} \sum_{i=1}^k c_{i, \lambda} \mathbf{z}(\frac{c}{\pi}X_i)^{\top} \overline{\mathbf{z}(\frac{c}{\pi}x)} = (\mathbf{y}^X)^{\top} \overline{\mathbf{z}(\frac{c}{\pi}x)},
\end{equation}
\noindent
where  $\mathbf{y}^X =  \frac{c}{n \pi} \sum_{i=1}^k c_{i, \lambda} \mathbf{z}(\frac{c}{\pi}X_i)$.
Note that the expansion coefficient vector $\mathbf{C}_{\lambda}$ for the VFA is also obtained with (\ref{eq:Tikhonov:expansion}) but using the Gram matrix calculated for similarities between FPEs $\mathbf{z}(\frac{c}{\pi}X_i)$.

\paragraph{\textit{\textbf{Empirical evaluation}}}\mbox{}\\ 
In order to demonstrate FPE-based methods for nonparametric nonlinear regression and their numeric counterparts, we used the same synthetic data as in~\citet{bousselmi2020reproducing} (cf. Example 1 therein): 
\noindent
\begin{equation}
    f(x)=\mbox{sinc}(\frac{20}{\pi}x), x \in I;
\end{equation}
\noindent
where $c=20$.
For the training set, values of $X_i$ were independent and uniformly distributed, while the values of $Y_i$ were obtained as:
\noindent
\begin{equation}
    Y_i=f(X_i) + \eta_i, \eta_i \in \mathcal{N}(0, 0.01).
\end{equation}
\noindent
Following the setup in~\citet{bousselmi2020reproducing}, the value of $c$ for the empirical projection method  was set to $20$ while for the Tikhonov regularization method the parameters were: $c=30$, $\lambda=0.01$.
\begin{figure}[H]
\centering
\includegraphics[width=1.0\columnwidth]{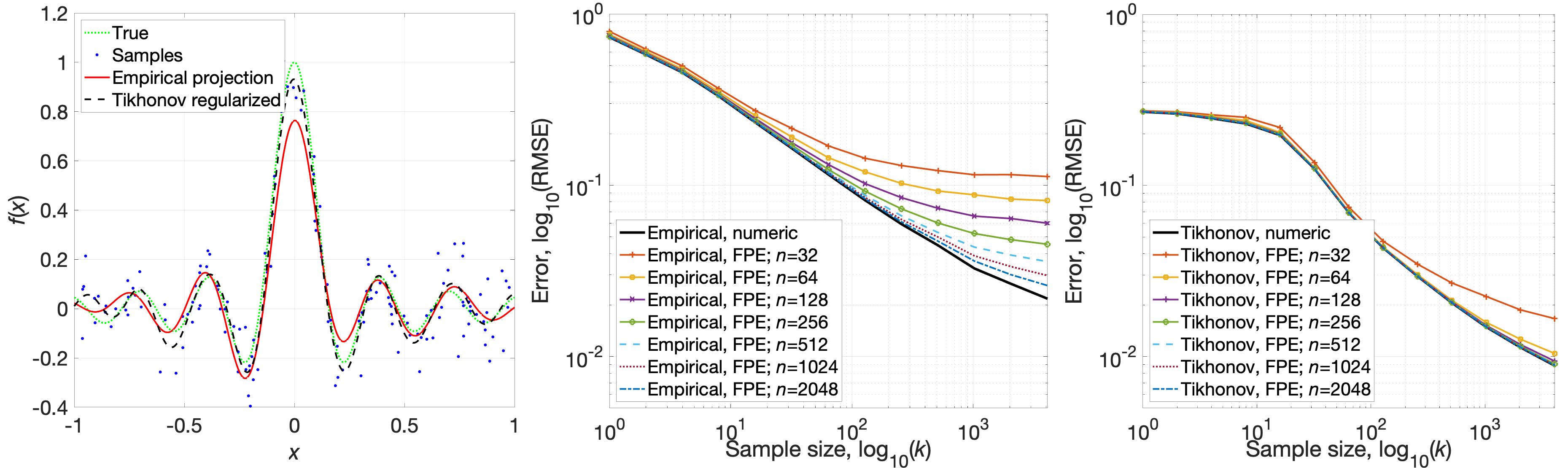}
\caption{
Nonlinear regression with VFA. 
Left panel shows the target function, $f(x)=\mbox{sinc}(\frac{20}{\pi}x)$, (green line), $k=150$ randomly drawn noisy samples from $f(x)$ (dots), and the estimations $\hat{f}(x))$ obtained with VFAs implementing empirical projection (orange line) and Tikhonov regularization (dashed black line) methods.
Middle and right panels depict the logarithms of the RMSE against the number of samples $k$ for empirical projection (middle panel) and Tikhonov regularization (right panel). The results are shown for numeric calculations of the kernel and for seven different dimensionalities of the VFA. 
The number of samples varied in the range $2^{[0,12]}$ with step $1$.
RMSE values are averages computed from $500$ simulation runs.
For each run, samples and noise added to $f(x)$  were drawn randomly, for the variants using the FPE, the base vector $\mathbf{z}$ was also chosen randomly at each run. 
}
\label{fig:regression:rmse}
\end{figure}

This experiment demonstrates that the VFA implementations of both regularization methods provide decent estimates of the target function (left panel) in Figure~\ref{fig:regression:rmse}.  
Compared to the empirical projection VFA, the Tikhonov VFA achieved lower RMSE. This improvement, however, comes at additional computational costs, as the Tikhonov VFA one needs to solve (\ref{eq:Tikhonov:expansion}).

Concerning the scaling of performance with number of available data points and VFA dimensionality, we observed similar trends as for density estimation. The error decreases with increased $k$ and $n$ and for fixed $k$ there is a minimum dimensionality that reaches optimal performance. Interestingly, the required VFA size for optimal performance is significantly smaller for the Tikhonov VFA than for the empirical projection VFA. 


\section{Discussion}
\label{sec:discussion}

\subsection{Summary of results}
Our main contribution is the formal characterization of vector function architectures (VFA), a new framework for computing on functions by manipulating high-dimensional randomized vectors. This work generalizes vector symbolic architecture (VSA) models for symbolic reasoning and it formalizes some of the earlier proposals to use VSA on subsymbolic data (see Section~\ref{prev_appli} below). 
Our results can be put into the following statements: 
\begin{itemize}
    \item The combination of VSA with a KLPE that is compatible with the VSA binding operation induces a Vector Function Architecture (VFA), i.e., a reproducing kernel Hilbert space (RKHS) in which functions are represented by vectors and can be manipulated by vector operations that form an algebra over the representation space. The algebraic vector operations can perform the following function operations: point-evaluation,  integration, addition, shift, and convolution (Theorem~1). 
    
    \item 
    Vector binding of two symbol representations in VSA forms a representation of the symbol pair, expressing, for example, a key-value or role-filler association between the symbols. It is interesting that in the framework of VSA, binding of function vectors expresses function convolution\footnote{This result should not be too surprising because the vector binding in VSA and VFA is implemented by convolution (circular convolution binding), or convolution in Fourier space (Hadamard binding).}.  Convolution has been proven a very important composition/decomposition operation for functions, as exemplified lately by the success of convolutional neural networks \citep{bengio2021deep}.

    \item Fractional Power Encoding (FPE), an existing LPE method based on the VSA binding operation, induces VFAs for one- and multi-dimensional data spaces. The kernel shape depends on the random structure of the FPE base vector. Uniformly sampled base vectors produce VFAs with a universal kernel shape, the sinc function, independent of the underlying binding operation. Thus, vectors in the resulting  VFA can represent the band-limited functions (Theorem~2). 
    
    \item
    A recipe is described how to shape the kernels in VFA. Specifically, any kernel with a Fourier density band-limited within $[-1/2, 1/2]$ can be produced in a VFA by sampling the FPE base vector accordingly. In particular, it is shown how to construct VFE with kernels useful to various applications, such as Gaussian kernels, non-Cartesian multi-dimensional sinc kernels, periodic grid kernels, etc.
    
    \item Like in traditional symbolic VSAs, VFAs require a mechanism for detecting, decoding and denoising of representations. A circuit mechanism is proposed to serve this purpose.
    
    \item We showcase a few examples of kernel computations with VFA, image processing, density estimation and nonlinear regression.

    \item Other types of LPEs (that have been used with VSA in the past), such as float codes, random vector concatenation, and random projections can also induce RKHS function spaces. 
    However, none of these other LPEs induce a VFA because they do not include a binding operation compatible with the encoding scheme, see Supplement~\ref{otherlpe}. 
\end{itemize}

\subsection{Technical applications of VFA}
In addition to VFA applications pioneered by earlier related work in Section~\ref{prev_appli} below, the exact definition of function space and algebraic function operations provided here pave the way for a host of novel applications. They include:

\begin{itemize}
    \item {\it VSAs/Hyperdimensional Computing}: VFA provides an algebraic, theoretically grounded approach to expand VSA/Hyperdimensional Computing techniques 
    to computing with functions. For an application example, see Section~\ref{sec:proc_im_dat}.
    
    \item {\it Large-scale kernel machines}: VFA provides feature representations and vector operations for implementing kernel methods at large scale. The idea to overcome the curse of dimensionality of traditional kernel methods \citep{bachall} by representing kernels as inner products of feature vectors has been pointed out earlier in \citep{rahimi2007random}, and see Section~\ref{ra_fea}. For examples of kernel applications of VFAs, see Section~\ref{sec:vfa_nonpara_ker_met}.
    
    \item {\it Probabilistic data structures/sketches}: Probabilistic data structures or sketches are reduced representations of mathematical objects formed from data points \citep{mitzenmacher2017probability}. 
    Such reduced representations can be more computation- and memory-efficient than storing all the individual data points. 
    In essence, a VFA vector can be seen as a compact probabilistic data structure or sketch of a function (Section~\ref{sec:vfa_nonpara_ker_met}) or an object (Section~\ref{sec:proc_im_dat}). 
    Interestingly, VFAs add new capabilities to probabilistic data structures. Specifically, a single VFA vector can implement the functionality of a kernel machine, going beyond the standard application domain of sketches, such as membership testing (Bloom filters~\citep{Bloom1970space}) or frequency estimation (count-min sketches~\citep{CountMin}) -- for an example, see Section~\ref{sec:vfa_nonpara_ker_met}.

    \item {\it Reservoir computing}: It has been emphasized previously that reservoir computing can implement the algebraic framework of VSAs in recurrent networks \citep{Frady2018,KleykointESN2020}. Changing the input encoding from random projection methods in conventional reservoir computing to KLPEs will yield a new class of recurrent networks for computing in a transparent fashion in a well-defined function space.
    
    \item {\it Neural networks}: 
    It has been posited that the missing of a binding operation in neural networks is the root of their lack of data efficiency, generalization, and robustness \citep{greff2020binding}. The generalization of the VSA binding concept from a symbolic domain to a function domain by VFAs is a critical step towards developing neural network approaches with a binding operation. Also, by representing functions as vectors, VFA provides an interesting input interface for processing functions with current-type neural networks. Further, there are known connections between kernels and deep learning, specifically, gradient-based learning can be formulated in terms of so-called path kernels \citep{domingos2020every}, an extension of neural tangent kernels \citep{jacot2018neural}. Whether or not, such kernels can be expressed in VFAs, is an open problem for future research.  
    
    \item {\it Dynamic cognitive modeling}: To model cognitive function of the brain, it has been suggested, e.g., \citep{port1995mind,eliasmith1996third}, to map discrete symbolic reasoning to continuous dynamical systems, specifically to dynamic neural field models \citep{amari1977dyn,ermentrout1993existence,jirsa1996field,erlhagen2002dynamic}. These models introduce a continuous, low-dimensional topological space as internal or ``mental'' navigation space. It has been emphasized how vector-symbolic concepts enable the construction of dynamic cognitive models but challenging inverse problems remain, i.e., how to appropriately design the mental space and a neural dynamics suitable to solve a given cognitive problem  \citep{beimgrabenpotthast2009inverse, WiddowsContinuous2015}. Using VFA enables the shaping of the similarity structure in the mental space, potentially offering new approaches to tackle the inverse problems of challenging cognitive tasks. 
    
\end{itemize}
 

\subsection{VFA models for neuroscience}

The predictions of VFA as a model for neuronal coding in the brain depend critically on the choice of the encoding/binding method.  The binding operations of classical VSAs, Hadamard product and circular convolution, do not easily map onto the function of biological neuronal circuits. They also require dense representation vectors that seem incompatible with the observation of sparse activity patterns in neural recordings \citep{RachkovskijBinding2001,frady2020variable}. 
Interestingly however, phasor vectors, as used in Hadamard FPE, can be naturally represented by spikes, where the complex phases are represented by the timing of periodic spike patterns \citep{frady2019robust}.
One can introduce population sparsity in phasor codes and build high-capacity associative memory networks for patterns with analog phase angles in a subset of neurons \citep{frady2019robust}.  Block-local circular convolution (Section~\ref{sec:blolocico}) is a binding operation that can operate with such codes. Potentially, this binding method could be implemented by biological mechanisms \citep{frady2020variable}, for example, leveraging active dendrites and coincidence detection \citep{schaefer2003coincidence}. 

There are previous attempts to build models for hippocampus/entorinal cortex using FPEs. 
One of these models with a spike-timing phase code exhibited phenomena that coincide with experimental observations, such as place fields and phase precession \citep{fradyframework18}.  Another model, based on rate codes and circular convolution binding \citep{KomerNavigation2020}, showed that the model neurons can exhibit grid-cell like responses. 
It is intriguing to speculate whether hippocampus and entorhinal cortex could implement and leverage full VFA functionality. 
A VFA model of hippocampus/entorhinal cortex would predict that activity patterns can represent functions over the environmental space, such as probability densities of rewards and future paths. Although the rate-based model implements a full VFA, the ability to represent functions was not exploited in the original publications. 
The phase code hippocampus model \citep{fradyframework18} could be extended to a full VFA by adding a binding operation like block-local circular convolution as described in Section~\ref{sec:blolocico}.    

VFA sheds new light on potential computational roles of neural coding. Neural coding is often seen as the outcome of statistical learning on the intrinsic structure of sensory signals. For example, source coding \citep{Shannon1949} or redundancy reduction \citep{barlow2001redundancy} are objectives that can often result in dimensionality reduction, but also in dimensionality expansion, when combined with sparsity \citep{Olshausen1996}.
Other theories for neural coding are based on random sampling, not learning, for example, using randomized synaptic projections. The computational objectives for such models include hashing for forming an expanded address space for improving signal detection \citep{babadi2014sparseness, fusi2016neurons, dasgupta2017neural}, or compressed sensing \citep{donoho2006compressed,candes2006compressive} for optimizing neural communication under wiring constraints \citep{isely2010deciphering, hillar2015can, ganguli2012compressed}. 
VFA reconciles learning-based and randomness-based theories for neural coding. A VFA theory makes the following predictions: a) Meaningful low-dimensional manifolds of sensory signals are extracted (not discussed here) and then re-coded by high-dimensional randomized activity patterns. The inner product in the high-dimensional representation space approximates a similarity kernel on the data manifold. 
b) The high-dimensional vector space can represent and manipulate points and functions on the data manifold. The similarity between points and available function space is determined by the kernel.  
c) Learning is approximation in the kernel function space. d) Binding between vectors enables the composition of new functions from already learned function primitives by convolution. e) The role of neural correlations and ``signal mixing'', as vaguely defined earlier, can be made explicit: Correlations in population activity encode information, which can be decoded by comparing the population activity vector with memorized vectors. Signals can be mixed two in different ways, a similarity-preserving manor by bundling, or in a similarity-destroying manor by binding. 

Most computational theories of brain function, such as Bayesian inference, predictive coding, etc., require the encoding of functions by the population activity of neurons.
Probably closest to the VFA concept are population codes \citep{pouget2000information, barber2003neural}, such as Bayesian population codes \citep{ma2006bayesian}. 
In these models each neuron typically has a Gaussian-shaped receptive field on the encoding manifold. This leads to an inner product kernel that decays with distance and is translation invariant.  Thus Bayesian population codes induce a kernel function space. 
However, they lack the binding operation (at least we are not aware of one) to perform the algebraic function manipulations possible with VFA.


\subsection{Related work}
Although we believe we are the first to formally define VFA models and describe their ability to represent and manipulate functions, there is related previous work. 

\subsubsection{Earlier applications of fractional power encoding in VSA}
\label{prev_appli}
Earlier proposals to combine VSA with FPE have pointed to a host of interesting applications. For example, the fractional power vector based on the circular convolution binding was proposed in~\citet{PlateRecurrent1992} as mechanism for representing discrete sequences in recurrent neural networks. Generalizing the fractional power vector, FPE based on circular convolution was proposed in a VSA for representing continuous trajectories in a 2-D space (Section 5.6 in~\citet{Plate1994}). 
The combination of VSA and circular convolution FPE\footnote{
In~\citet{KomerContinuous2019}, the model combining these elements is referred to as ``spatial semantic pointers'', extending the convention of the same research group to refer to VSA representations as ``semantic pointers''~\citep{BlouwConcepts2016}.  
}
was revisited in a number of more recent papers in the context of the following applications:
\begin{itemize}
    \item {\it Reasoning on 2-D images}: \citet{WeissOlshausenSpatial16} used this model to holistically represent 2-D images, thus offering the possibility to query the images, i.e. answer relational queries such as ``which digit is below a 2 and to the left of a 1?'' in images containing an array of MNIST digits. Similar models for reasoning on images were also described in~\citet{FradyDisentangling2018, LuFractional2019}. Our example in this application domain demonstrates how the kernel shapes of a VFA can be adapted to optimally accommodate this application, i.e., shaping a periodic kernel for providing toroid boundary conditions, see Section~\ref{sec:proc_im_dat}.
    \item {\it Navigating a 2-D environment}:  \citet{WeissOlshausenSpatial16} have demonstrated the use of FPE in solving navigation problems, an application further elaborated in~\citet{KomerNavigation2020}. 
    
    \item {\it Neuromorphic computing models}: There are some initial efforts to use such models in neuromorphic computing. In~\citet{fradyframework18}, a Hadamard FPE was used in a hippocampus model linking computations and rhythm-based timing patterns in spiking neural networks, while~\citet{DumontGrid2020} proposed a way to implement FPE vectors by the rates of spiking neurons. 
    \item {\it Prediction of dynamical systems}: \citet{VoelkerFPEDynamical2021} have proposed to use the model for simulating and predicting the behavior of dynamical systems.
\end{itemize} 

\subsubsection{Kernel machines with random Fourier features}
\label{ra_fea}
It has been known that positive definite kernels are equivalent to inner products in a derived feature space \citep{scholkopf2002learning}.  \citet{rahimi2007random} pointed out how the Bochner theorem (Section~\ref{funct_ana_background}) allows to construct representation spaces with random Fourier features so that the inner product approximates a similarity kernel of a certain kind. Every Fourier density $p({\boldsymbol \omega})$ in (\ref{bochner}) defines such a similarity kernel. 
If the Fourier density is
$[-1/2,1/2]$-band-limited, it induces a Hadamard FPE:
\begin{equation}
    \mathbf{z}(\mathbf{x}) = \mathbf{z}_1(x_1) \odot \mathbf{z}_2(x_2) \odot \cdots \odot \mathbf{z}_m(x_m)
    \label{rahimi_bochner_encode}
\end{equation}
with the base vectors being random phasors $\mathbf{z}_i(1) = \left(e^{\imath \phi^i_1},..., e^{\imath \phi^i_n} \right)$ sampled from $p({\boldsymbol \phi} = {2 \pi \boldsymbol \omega})$.  
Thus, random Fourier features contain Hadamard FPEs as a special case. 
Conversely, FPEs with binding operations other than the Hadamard product constitute encoding schemes that are not contained in random Fourier features. 
Although random Fourier features enable Hadamard VFAs, to our knowledge this potential has not been exploited. Specifically, \citet{rahimi2007random} did not use Hadamard binding between representation vectors to manipulate functions as described by Theorem~1.


\subsubsection{The discrete Fourier transform}

The Hadamard FPE generalizes another simple construction of an LPE that comes to mind, the representation of points by the discrete Fourier transform of the band-limited delta function, the sinc function. First a finite interval of the (one-dimensional) data space is tiled equally by $n$ points, $x_0,...,x_{n-1}$. The representation of an arbitrary point $r$ is then given by:
\begin{eqnarray}
    \mathbf{z}(r) &=& {\cal F}\left(\sqrt{\frac{2}{\pi}}\mbox{sinc}(x+r)\right) = 
        \int d\omega  e^{ \imath \omega r} \Pi\left(\frac{\omega}{2}\right) \nonumber\\
            &=& F (\mbox{sinc}(x_0+r), \mbox{sinc}(x_{1}+r), \cdots, \mbox{sinc}(x_{n-1}+r) )^{\top}\nonumber\\
    &=& \left( (e^{\imath \omega(-1/2)}, e^{\imath \omega(-1/2+1/n)}, \cdots, e^{\imath \omega(-1/2+ (n-1)/n)})^r \right)^{\top}
    \label{bl_delta_function_encoding}
\end{eqnarray}
with $\Pi(x)$ the box distribution. Thus, (\ref{bl_delta_function_encoding}) is a Hadamard product FPE with a particular base vector, which is a regular phasor vector, not a random phasor. The base vector is essentially a column of the DFT matrix $F$. The important generalization provided by our treatment here is that every random phasor seed induces an LPE, and for pseudo-orthogonal random seeds the corresponding LPEs do not interact (or, more correctly, only interact with very low probability), and thus, can be combined to represent multi-dimensional spaces.

\bibliographystyle{apalike}
\bibliography{references,hdspike,capacity}

\newpage

\appendix

\begin{center}
    {\Huge\bf Supplemental material}
\end{center}

\section{Multi-dimensional Hadamard FPE}
\label{sec:mdim_hadamard-fpe}
In Section~\ref{sec:multidimensional:FPEs} we describe how the binding operation can be used to construct LPEs for multi-dimensional FPE from individual one-dimensional FPEs. Here 
we describe and analyze in more detail the Hadamard FPE for multi-dimensional data. 
The following  interactions between inner product and Hadamard binding simplify the notation:
\begin{eqnarray}
\left[\mathbf{x}\odot \mathbf{y}\right]^{\top} \overline{\mathbf{u}\odot \mathbf{v}} 
    &=& \sum_i x_i y_i \overline{u_i} \overline{v_i}\label{innerhadamard}\\
    \left[\odot_{j=1}^m \mathbf{x}^j\right]^{\top} \overline{\left[\odot_{j'=1}^{m} \mathbf{y}^{j'}\right]} 
    &=& \sum_i \prod_{j=1}^m x^j_i\overline{y^{j}_i} \label{innerhadamard_m}
\end{eqnarray}


In a Cartesian multi-dimensional Hadamard FPE, individual data dimensions are encoded by one-dimensional FPEs and then combined via the binding operations: 
\begin{eqnarray}
    \mathbf{z}(\mathbf{x}) &=& \odot_{j=1}^m \mathbf{z}^j(x_j)
    \label{high-dime_hadamard_fpe}\\
    \mathbf{z}^j(x_j) &=&  (^j\mathbf{z})^{x_j} = (e^{\imath x_j \phi^j_1}, ..., e^{\imath x_j \phi^j_n})
    \label{one-dim_hadamard_fpe}
\end{eqnarray}
with base vectors of the one-dimensional FPEs $^j\mathbf{z} = (e^{\imath \phi^j_1}, ..., e^{\imath \phi^j_n})$ chosen independently for all $j$.
The following theorem shows that this FPE induces a VFA.

\vspace{0.5cm}
\noindent
{\it Theorem 3:} Using Hadamard binding to construct a multi-dimensional LPE encoding method from one-dimensional Hadamard FPEs (\ref{high-dime_hadamard_fpe}) yields an inner product kernel:
\begin{equation}
    \mathbf{z}(\mathbf{x})^{\top} \overline{\mathbf{z}(\mathbf{y})} 
    = \sum_{i=1}^n e^{\imath \sum_{j=1}^m \phi^j_i (x_j - y_j)} \to K_{\odot}(\mathbf{x}-\mathbf{y}), 
    \label{high-dime_hadamard_fpe:tens}
\end{equation}
which is translation invariant and positive definite. 
Thus, the multi-dimensional Hadamard FPE induces a VFA, representing functions with multi-dimensional domains, even if the dimension $n$ is not large enough so that the inner product closely approximates the multi-dimensional sinc kernel.

\vspace{0.5cm}
\noindent
{\it Proof:}
Using (\ref{innerhadamard_m}) and (\ref{one-dim_hadamard_fpe}), the inner product between representations of data points can be written:
\begin{equation}
     \mathbf{z}(\mathbf{x})^{\top} \overline{\mathbf{z}(\mathbf{y})} = \sum_{i=1}^n \prod_{j=1}^m  (^jz_i)^{x_j}\;\overline{(^jz_i)^{y_j}} 
    = \sum_{i=1}^n \prod_{j=1}^m  (^jz_i)^{x_j-y_j}
    = \sum_{i=1}^n e^{\imath \sum_{j=1}^m \phi^j_i (x_j - y_j)}
\end{equation}
An element of the Gram matrix between two data points $\mathbf{x}^l$ and $\mathbf{x}^{l'}$ 
can be transformed as follows:
\begin{eqnarray}
    M_{ll'}=
    \mathbf{z}(\mathbf{x}^l)^{\top} \overline{\mathbf{z}(\mathbf{x}^{l'})}
    = \sum_{i=1}^n  e^{\imath \sum_{j=1}^m \phi^j_i (x^l_j - x^{l'}_j)} 
    = \sum_{i=1}^n \overline{e^{\imath \sum_{j=1}^m \phi^j_i (x^{l'}_j - x^{l}_j)}},
    = \overline{M_{l'\;l}}
    \label{ishermitian}
\end{eqnarray}
which shows
that the Gram matrix is Hermitian, 
an, thus, its eigenvalues are real-valued. As for inner product kernels in general, the Gram matrix is positive definite \citep{hofmann2008kernel}.  Because of these conditions, the multi-dimensional Hadamard LPE induces a function space.

\noindent
$\Box$\\

In the resulting VFA, the vector representation of functions over the multi-dimensional data space (\ref{mult_function_rep}) is given by:
\begin{eqnarray}
    (y_f)_i &=& \sum_k \frac{\alpha_k}{n} z^1_i(r^k_1) \; z^2_i(r^k_2) \;...\;  z^m_i(r^k_m)
    = \sum_k \frac{\alpha_k}{n} e^{\imath \sum_{j=1}^m r^k_j \phi^j_i}
\end{eqnarray}
and function evaluation at a point $\mathbf{s}$ is:
\begin{equation}
    f(\mathbf{s}) 
    = (\mathbf{y}_f)^{\top} \overline{\mathbf{z}(\mathbf{s})}
    = \sum_{k} \frac{\alpha_k}{n} \sum_{i=1}^{n} \prod_{j=1}^m z^j_i(r^k_j)\overline{z^j_i(s_j)}
    = \sum_k \frac{\alpha_k}{n} \sum_{i=1}^{n} e^{\imath \sum_{j=1}^m (r^k_j-s_j) \phi^j_i}.
\end{equation}

One might question that (\ref{high-dime_hadamard_fpe}) is the best way to form a multi-dimensional FPE from combining one-dimensional FPEs. As known from classical VSAs, Hadamard binding provides a reduced, lossy representation of tensor product binding \citep{frady2020variable}. To avoid this loss, one could alternatively 
combine the one-dimensional FPEs with Smolensky's tensor product \citep{Smolensky1990}. For $m=2$ input dimensions, the representation tensor is the outer-product matrix:
\begin{equation}
    \mathbf{Z}(\mathbf{x}) = \mathbf{z}^1(x_1) \overline{\mathbf{z}^2(x_2)}^{\top}
                           = (^1\mathbf{z})^{x_1} \overline{(^2\mathbf{z})^{x_2}}^{\top}
    \label{smolensky_hadamard_fpe}
\end{equation}
Here the Frobenius inner product of tensors (Hadamard product of the matrices and addition of all matrix elements) converges to the exact Cartesian sinc kernel:
\begin{equation}
    \mathbf{Z}(\mathbf{x})^{\top} \overline{\mathbf{Z}(\mathbf{y})} \to K_{sinc}(x_1 - y_1) \times K_{sinc}(x_2 - y_2)
\label{eq:frobenius_inner_product}
\end{equation}
This construction with the tensor product does indeed accelerate the convergence to the ideal 2-D sinc kernel with increasing $n$, see Figure~\ref{fig:2d_cartesian_sinc}. However, this improvement comes at the expense of requiring for the representation of 2-D points a dimension of $n^2$, whereas with the Hadamard product the representations of multi-dimensional points are always just $n$. 
Figure~\ref{fig:2d_cartesian_sinc}D compares the two methods for 2-D points represented in a vector space of the same dimension. This comparison shows that representations formed by Hadamard product approximate the sinc kernel at nearly the same rate as the tensor product.

\section{Other methods for generating hexagonal kernels}
\label{sec:hex:other}

\cite{KomerNavigation2020} propose an alternative construction of kernels with hexagonal symmetry. They do so by projecting a 3-D Cartesian kernel to form a 2-D hexagonal grid. The 3-D Cartesian grid uses three random unitary vectors $z_1,z_2,z_3$. The coordinate in a 2-D space $\textbf{x} = (x,y)$ is then represented by the vector
\begin{equation}
\mathbf{z}(\mathbf{x}) = \mathbf{z}_1^{cc}(\mathbf{\xi}_1^\top \mathbf{x}) \circledast  \mathbf{z}_2^{cc}(\mathbf{\xi}_2^\top \mathbf{x}) \circledast  \mathbf{z}_3^{cc}(\mathbf{\xi}_3^\top \mathbf{x})
\label{komerformula_cc}
\end{equation}
where ${\xi_1,\xi_2,\xi_3}$ are the hexagonal basis vectors defined for equation (\ref{eq:hexkern}). Though \cite{KomerNavigation2020} opt for unitary vectors with binding via circular convolution, in principle other types of VFA could also be used.

This 3-D system can be reconfigured to be a 2-D system by forming base phasor vectors that do not have independently uniform phases. We can form the vectors $\mathbf{z}_x = \mathbf{z}_1^{cc}(\mathbf{\xi}_1[0]) \circledast \mathbf{z}_2^{cc}(\mathbf{\xi}_2[0]) \circledast \mathbf{z}_3^{cc}(\mathbf{\xi}_3[0])$, and $\mathbf{z}_y = \mathbf{z}_1^{cc}(\mathbf{\xi}_1[1]) \circledast \mathbf{z}_2^{cc}(\mathbf{\xi}_2[1]) \circledast \mathbf{z}_3(\mathbf{\xi}_3^{cc}[1])$. Now the coordinate $(r, s)$ is expressed naturally by
\begin{equation}
  \mathbf{z}(r,s) = \mathbf{z}_x^{cc}(r) \circledast \mathbf{z}_y^{cc}(s)  
\label{twobasehexagon}  
\end{equation} 

\begin{figure}[H]
\centering
\includegraphics[width=0.7\columnwidth]{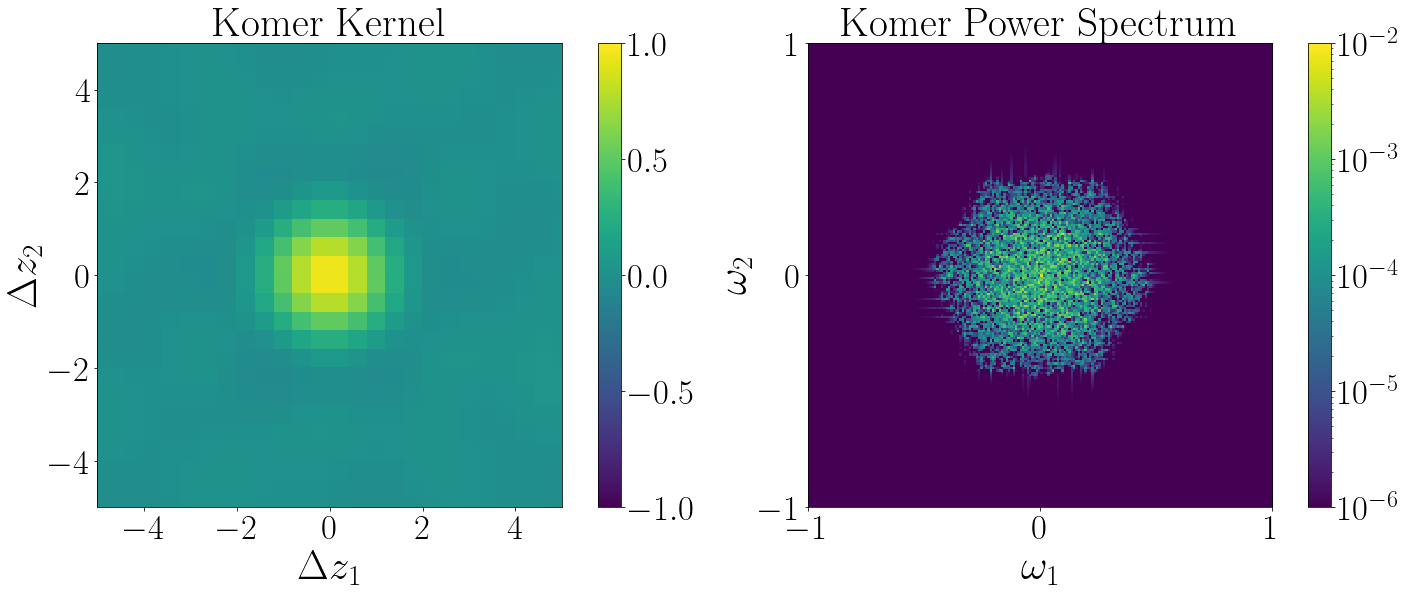}
\caption{
Example of kernel generated using the method of \cite{KomerNavigation2020} and its respective power spectrum ($n=5,000$). (Power spectrum is normalized such that total power equals 1.)
}
\label{fig:komer_kernel}
\end{figure}

Note that although this kernel exhibits hexagonal symmetry, it does not converge to the hexagonal sinc function kernel. Thus, it is not equivalent to  (\ref{concatenatedprojections}), which does converge to the sinc function for large dimensionality. Figure \ref{fig:komer_kernel} shows some of the differences. In particular, while the power spectrum is roughly uniform within the spectra of the hexagonal sinc function kernel, for  (\ref{komerformula_cc}) the lower frequencies have higher power.

\section{Other types of LPE and their kernels}
\label{otherlpe}

In order to process subsymbolic data in VSA, such as in machine learning problems \citep{RachkovskijClassifiers2007, Rasanen2015tr,KleykoIndustrial2018,RahimiBiosignal2019,schindler2021primer}, previous work combined VSAs with different kinds of LPEs for representing subsymbolic data with vectors.
Here we briefly review the LPE methods that have been used in the past and assess whether or not they induce a VFA. The encoding methods use either real (Section~\ref{sec:LPE:RP}) or binary/bipolar (elsewhere in this section) vector spaces.

\subsection{The thermometer code}

\begin{figure}[H]
\centering
\includegraphics[width=0.8\columnwidth]{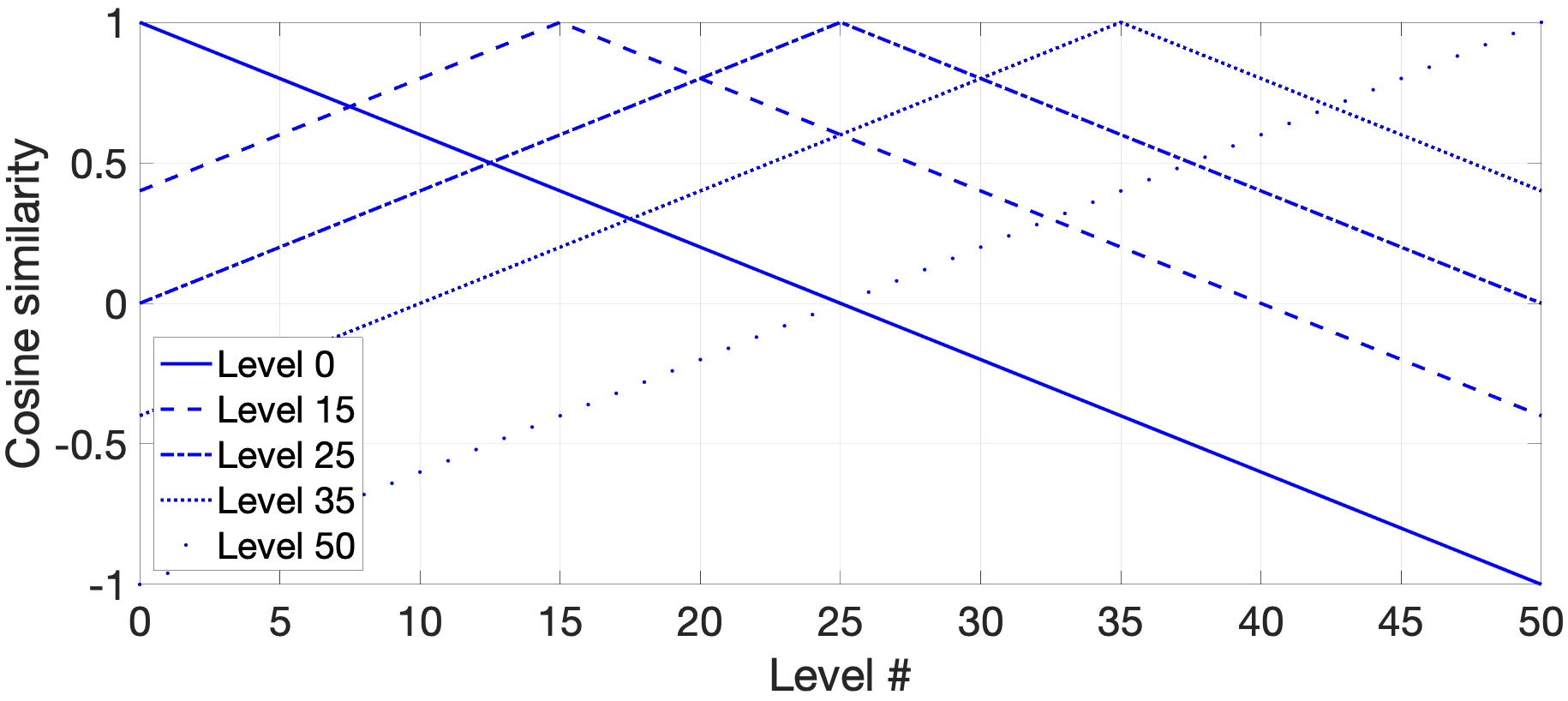}
\caption{
Similarity kernel of the thermometer code shown for several levels; $n$ was set to $50$.
}
\label{fig:thermometer}
\end{figure}

The thermometer code~\citep{PenzCloseness1987, RachkovskijScalars2005, BuckmanThermometer2018, KleykoDensityEncoding2020} is a simple and structured way to form an LPE for a range of discrete levels $s$, $s \in [0,n]$. 
The first code $\mathbf{z}(0)$ consists of all -1s. 
For other levels the components of $\mathbf{z}(s)$ are determined as:
\begin{equation}
    z_i(s)= 
\begin{cases}
    +1,& i\leq s\\
    -1,              & \text{otherwise}
\end{cases}
\end{equation}
Thus, the last code $\mathbf{z}(n)$ consists of all +1s.
So in total the thermometer code can represent $n+1$ levels. 
Figure~\ref{fig:thermometer} shows how the cosine similarity looks like for several different levels when $n=50$. Thermometer codes produce a translation-invariant kernel which is triangular and has a width of $2n+1$ levels. It is thus a nonlocal kernel, that is, there are no two points in the encoding range that have a similarity of zero. In practice, thermometer codes are used commonly when applying VSAs to classification problems.

\subsection{The float code}

\begin{figure}[H]
\centering
\includegraphics[width=0.8\columnwidth]{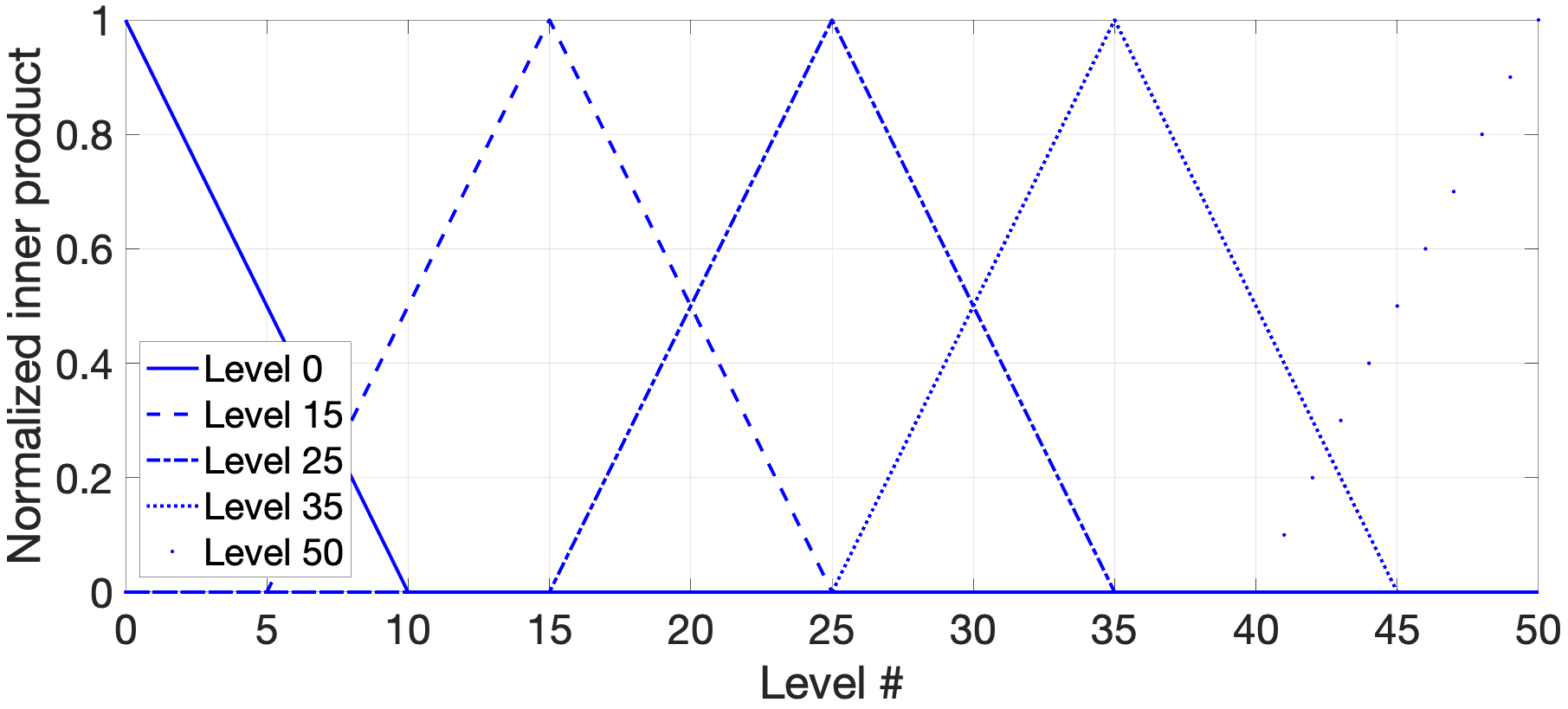}
\caption{
Similarity kernel of the float code shown for several levels; $n$ was set to $60$ while $w$ was set to $10$.
}
\label{fig:float}
\end{figure}

The float code, also known as the sliding code,~\citep{GoltsevFloat1996, RachkovskijScalars2005} addresses the issue of the thermometer code, i.e., that the similarity decay is not local. 
This is done by using $w$ consecutive +1 components (``float'') where the size of $w$ regulates similarity characteristics of the code. 
For the binary case, the similarity kernel of the float code is the triangular kernel of width $2w+1$ levels. 
To encode the lowest value $\mathbf{z}(0)$, the first $w$ components of the vector are set to +1s while the rest of the components are 0s.
In general, the components of $\mathbf{z}(s)$ are determined as:
\noindent
\begin{equation}
    z_i(s)= 
\begin{cases}
    +1, & s\leq i<s+w\\
    0,              & \text{otherwise}
\end{cases}
\end{equation}

Figure~\ref{fig:float} depicts how similarity (inner product normalized by $w$) decays for several levels in the float code for $n=60$, $w=10$. The float code also produces a triangular kernel, but
in contrast to the thermometer code, it allows controlling the width of the triangular kernel. 
The number of levels it could encode is still limited and equals to $n-w+1$. 
Thus, the float code induces an RKHS which can represent functions composed from its triangular kernel, but it cannot induce a VFA since there is no VSA binding operation known that the float code would be compatible with (\ref{transbybind}).

\subsection{Encoding by concatenation of parts of reference vectors}

\begin{figure}[H]
\centering
\includegraphics[width=0.8\columnwidth]{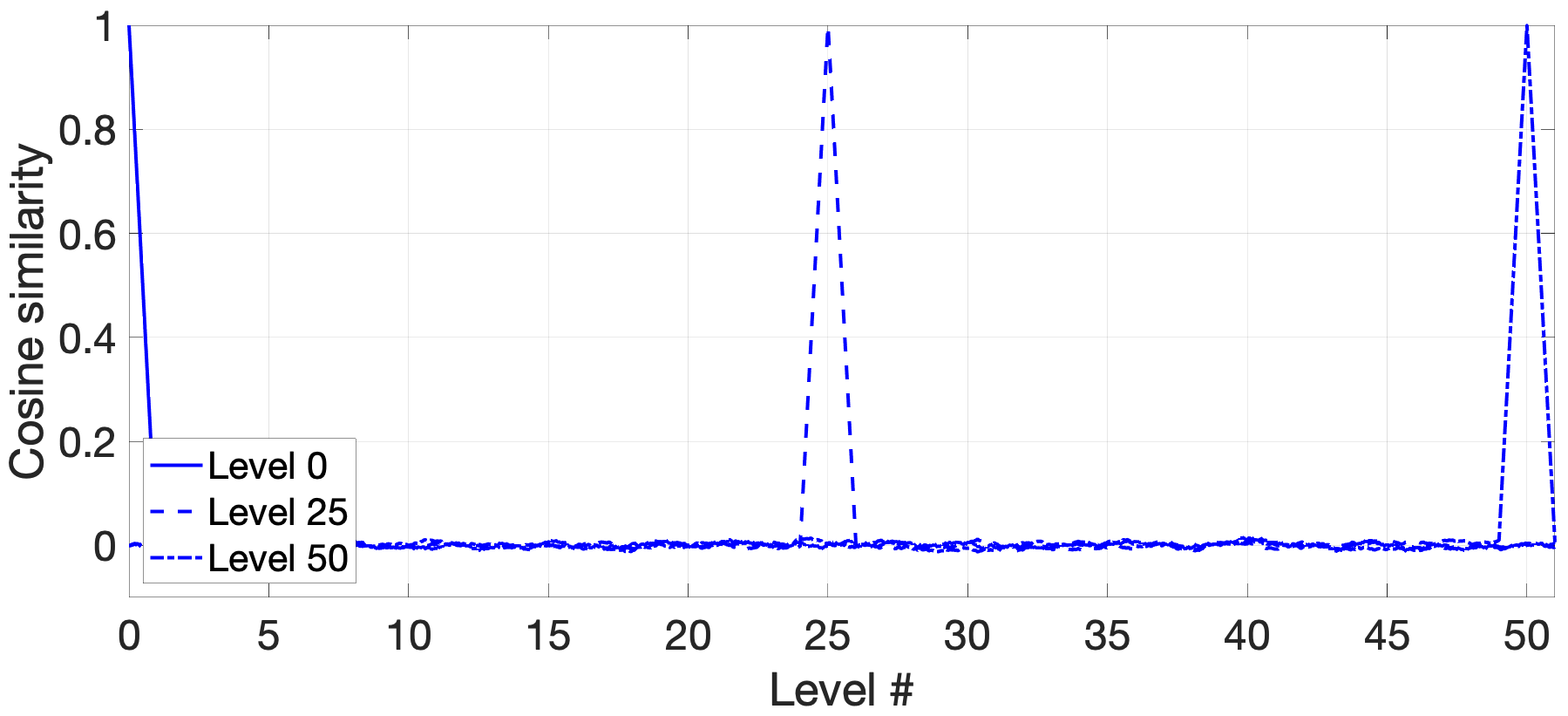}
\caption{
Similarity kernel for the encoding by concatenation;
$n$ was set to $1000$.
The values of similarities were averaged over $50$ random initializations of the code. 
}
\label{fig:concatenation}
\end{figure}

When encoding by concatenation of parts of reference vectors~\citep{RachkovskijScalars2005}, we have to choose some ``anchor'' levels (e.g., integers in the range $[0,100]$).
These ``anchor'' levels are assigned with randomly chosen vectors. 
Values $x$ between two ``anchor'' levels $s$ and $s+1$ ($s<x<s+1$) are represented by concatenating the proportional amount from the corresponding vectors $\mathbf{z}(s)$ and $\mathbf{z}(s+1)$:
\noindent
\begin{equation}
    z_i(x)= 
\begin{cases}
    z_i(s+1),& i \leq  \lfloor  (x-s)n \rceil  \\
    z_i(s),              & \text{otherwise}
\end{cases}
\end{equation}

Figure~\ref{fig:concatenation} shows how the cosine similarity looks for several different ``anchor'' levels. 
As we can see, the similarity within two neighboring ``anchor'' levels decays according to the triangular kernel.
The advantage of the encoding by concatenation is that the number of ``anchor'' levels can be very large even for relatively small $n$ as ``anchor'' levels are represented by random vectors, however, the disadvantage is that the similarity is only preserved between nearby ``anchor'' levels. As for the float code, the concatenation method induces the RKHS with trianglular kernels but it does not induce VFA due to the absence of the proper binding operation to satisfy (\ref{transbybind}).

\subsection{Scatter code}

\begin{figure}[H]
\centering
\includegraphics[width=0.8\columnwidth]{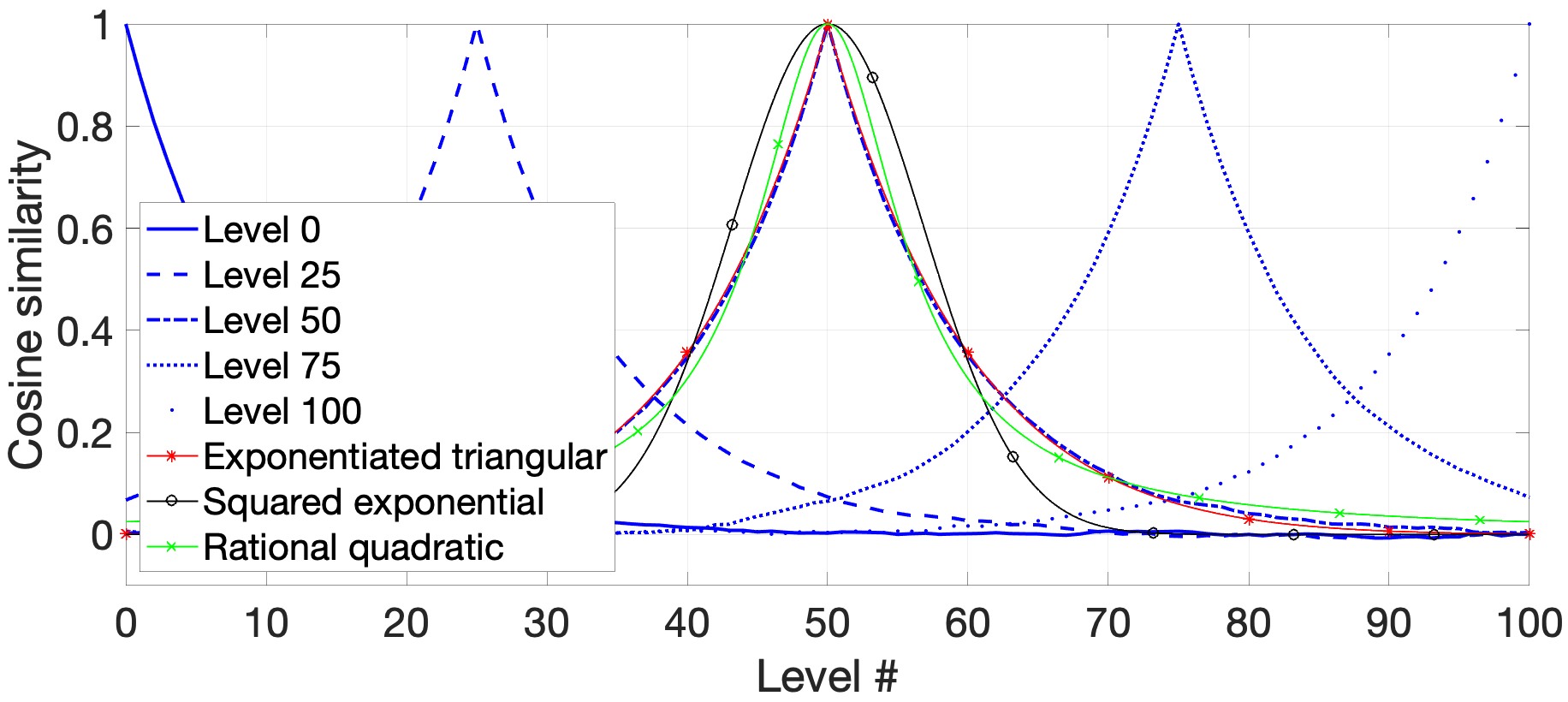}
\caption{
Similarity kernel of a scatter code; $n$ was set to $1000$, $p$ was $0.05$.
The values of similarities were averaged over $50$ random initializations of the code. 
}
\label{fig:scatter}
\end{figure}

Scatter codes~\citep{SmithScatter1990, RachkovskijScalars2005, kleyko2018classification} are another alternative to form an LPE where similarity decays nonlinearly.  
In scatter codes, the code for the first level $\mathbf{z}(0)$ is chosen randomly while each subsequent code is obtained from the previous one by randomly swapping its components with some probability $p$:
\begin{equation}
    z_i(s)= 
\begin{cases}
    -z_i(s-1),& r_i \leq  p  \\
    z_i(s-1),              & \text{otherwise}
\end{cases}
\end{equation}
where $r_i$ is a random value for $i$th component of $\mathbf{z}(s)$ chosen from the uniform distribution. 
Note that potentially there is no limitation on how many levels can be created with the scatter codes.

Figure~\ref{fig:scatter} shows how the cosine similarity looks for several different levels formed with the scatter code. Interestingly, the kernels are `bell-shaped'' with the exact shape depending on the parameter settings. 
To better figure out which standard kernel will correspond to this similarity, we have empirically fitted three kernels: exponentiated triangular: 
\begin{equation}
    K(s_1, s_2)= (1- \gamma |s_2-s_1|)^\alpha;
\end{equation}
squared exponential:
\begin{equation}
     K(s_1, s_2)= e^{-\frac{(s_2-s_1)^2}{2l^2}};
\end{equation}
 and rational quadratic:
\begin{equation}
     K(s_1, s_2)= \left(1+\frac{(s_2-s_1)^2}{2\alpha l^2} \right)^{-\alpha} ;
\end{equation} 
The parameters of the kernels were chosen using the mean squared error as the fit criterion. 
Figure~\ref{fig:scatter} shows that 
the scatter code can induce an RKHS with kernels that are approximately exponentiated triangular but similar to all previous LPEs there is no known VSA binding operation the scatter code would be compatible with (\ref{transbybind}). 

\subsection{Random projection encoding}
\label{sec:LPE:RP}


For a $d$-dimensional vector of real-valued input data $\mathbf{r}$, the random projection (RP) encoding is given by:
\begin{equation}
    \mathbf{z}(\mathbf{r}) = g(\mathbf{M} \mathbf{r}) 
\end{equation}
where $\mathbf{M} \in [n \times d]$ is a random matrix with its valued being drawn from, e.g., normal distribution and $g(\mathbf{x})$ a point-wise nonlinearity (e.g., $\mbox{sign}(\mathbf{x})$ can be used).
Such a combination of a linear RP, followed by a (often sparsifying) nonlinearity, have been proposed in many different contexts, for example, to 
activate the locations 
in an associative memory model \citep{kanerva1988sparse}, for neuro-biology inspired olfactory encoding \citep{dasgupta2017neural}, and for LPE in VSAs \citep{RachkovskijRP2012,thomas2020theoretical}. 
Note that without nonlinearity, the inner product of the representations is the same as the inner product between the data vectors:
\noindent
\begin{equation}
    K(\mathbf{r}_1, \mathbf{r}_2)= (n/2)(|\mathbf{r}_1|_2^2 + |\mathbf{r}_2|_2^2 - |\mathbf{r}_1-\mathbf{r}_2|_2^2 ),
\end{equation}
\noindent
where $|\mathbf{r}|_2^2$ is the squared L2 norm of vector $\mathbf{r}$. 
For such encoding functions, which preserve the inner product, the resulting kernel is not translation-invariant as it depends not only on the difference vector but on the individual vectors' norms. 
It is, however, possible to from translation-invariant kernels with RP if proper point-wise nonlinearity is used. 
As it was shown in \citep{rahimi2007random}, the use of cosine function and proper distribution of values of $\mathbf{M}$ will lead to distributed representations corresponding forming inner product kernels:
\noindent
\begin{equation}
    \mathbf{z}(\mathbf{r}) =  \cos (\mathbf{M} \mathbf{r} + \mathbf{b}),
\end{equation}
\noindent
where values of $\mathbf{b}$ are chosen uniformly from $[-\pi,\pi]$ while values of $\mathbf{M}$ are chosen from the Fourier transform of the desired kernel in the same way as described in Section~\ref{sec:shapingFPEkernels}.
However, we are not aware of any VSA binding operation that would make RP compatible with (\ref{transbybind}).





\end{document}